\documentclass[sn-mathphys-num]{sn-jnl}

\usepackage[numbers,sort&compress]{natbib}
\usepackage{graphicx}%
\usepackage{multirow}%
\usepackage{amsmath,amssymb,amsfonts}%
\usepackage{amsthm}%
\usepackage{mathrsfs}%
\usepackage[title]{appendix}%
\usepackage{xcolor}%
\usepackage{textcomp}%
\usepackage{manyfoot}%
\usepackage{booktabs}%
\usepackage{algorithm}%
\usepackage{algorithmicx}%
\usepackage{algpseudocode}%
\usepackage{listings}%
\usepackage{hyperref}
\usepackage{subfigure}
\usepackage{tabularray}

\theoremstyle{thmstyleone}%
\theoremstyle{thmstyletwo}%

\theoremstyle{thmstylethree}%

\begin{document}

\title[Article Title]{STAHGNet: Modeling Hybrid-grained Heterogenous Dependency Efficiently for Traffic Prediction}


\author[1]{\fnm{Jiyao} \sur{Wang}}

\author[2]{\fnm{Zehua} \sur{Peng}}

\author[4]{\fnm{Yijia} \sur{Zhang}}

\author*[1,4]{\fnm{Dengbo} \sur{He}}\email{dengbohe@hkust-gz.edu.cn}
\author[3,4]{\fnm{Lei} \sur{Chen}}


\affil*[1]{\orgdiv{Systems Hub}, \orgname{The Hong Kong University of Science and Technology (Guangzhou)}, \orgaddress{ \city{Guangzhou}, \postcode{511400 }, \country{China}}}

\affil[2]{\orgdiv{Computer Science College}, \orgname{Sichuan University}, \orgaddress{\city{Chengdu}, \postcode{610065}, \country{China}}}

\affil[3]{\orgdiv{Information Hub}, \orgname{The Hong Kong University of Science and Technology (Guangzhou)}, \orgaddress{\street{Street}, \city{Guangzhou}, \postcode{511400}, \country{China}}}

\affil[4]{\orgdiv{School of Engineering}, \orgname{The Hong Kong University of Science and
Technology}, \orgaddress{\city{Hong Kong S.A.R}, \postcode{999077}, \country{China}}}


\abstract{Traffic flow prediction plays a critical role in the intelligent transportation system, and it is also a challenging task because of the underlying complex Spatio-temporal patterns and heterogeneities evolving across time. However, most present works mostly concentrate on solely capturing Spatial-temporal dependency or extracting implicit similarity graphs, but the hybrid-granularity evolution is ignored in their modeling process. In this paper, we proposed a novel data-driven end-to-end framework, named Spatio-Temporal Aware Hybrid Graph Network (STAHGNet), to couple the hybrid-grained heterogeneous correlations in series simultaneously through an elaborately Hybrid Graph Attention Module (HGAT) and Coarse-granularity Temporal Graph (CTG) generator. Furthermore, an automotive feature engineering with domain knowledge and a random neighbor sampling strategy is utilized to improve efficiency and reduce computational complexity. The MAE, RMSE, and MAPE are used for evaluation metrics. Tested on four real-life datasets, our proposal outperforms eight classical baselines and four state-of-the-art (SOTA) methods (e.g., MAE 14.82 on PeMSD3; MAE 18.92 on PeMSD4). Besides, extensive experiments and visualizations verify the effectiveness of each component in STAHGNet. In terms of computational cost, STAHGNet saves at least four times the space compared to the previous SOTA models. The proposed model will be beneficial for more efficient TFP as well as intelligent transport system construction. }




\maketitle

\section{Introduction}\label{sec1}

Traffic Flow Prediction (TFP) is a foundational component of intelligent transportation systems (ITS), aiming to estimate the future traffic conditions of a designated location in a transportation network based on the historical value of flow sensor readings in a complex interactive environment. It is fundamental for the stability and safety of intelligent transportation systems \cite{yang2021unsupervised}, however, which is still challenging due to the complex transportation interactions and ubiquitous noise/perturbations in data.

There has been a lot of research on accurate TFP to tackle these challenges recently. Most traditional methods are usually a statistical model for time series forecasting, which can be divided into univariant time series forecasting and multivariate time series forecasting. Univariate time series learning methods \cite{1,2,3} mainly focus on the temporal correlations of the traffic flow time series from a single sensor. However, the information is not only the time series received by the sensor but also needs to consider the geographical location of the sensor in the whole transportation network, because the flow condition of the road impacts other roads. Other multivariate time series forecasting methods \cite{6,9} were proposed to identify the hidden spatial relationship between sensors at different times and apply it in time series prediction. 

In short-term time series data, traditional prediction models work relatively well, but do not have sufficient accuracy for long-term time series data. In general, the traffic flow data usually have a very long-term time dependence. For example, a section of road is congested on a certain day, and people stuck in traffic may not drive on this road at the corresponding time for a few days or a few months, so the short time series does not contain enough information to predict. As the amount of data, we need to predict increases gradually, the data becomes more and more complex, and the nonlinear characteristics of the time series in the data are more obvious, which leads to the insufficiency of the capacity of traditional methods. The recent deep-based method \cite{gu2018recent,schuster1997bidirectional,vaswani2017attention} can iteratively learn the intra- and inter-time-series temporal dependencies between multivariate sensors for TFP. Specifically, to represent spatial dependency of multivariate time series with the non-Euclidean spatial structure that is suitable for road network \citep{yu2017spatio, zhao2019t}, graph neural network (GNN) \citep{scarselli2008graph} is introduced to TFP. 

\begin{figure}\centering

\includegraphics[scale=0.4]{ 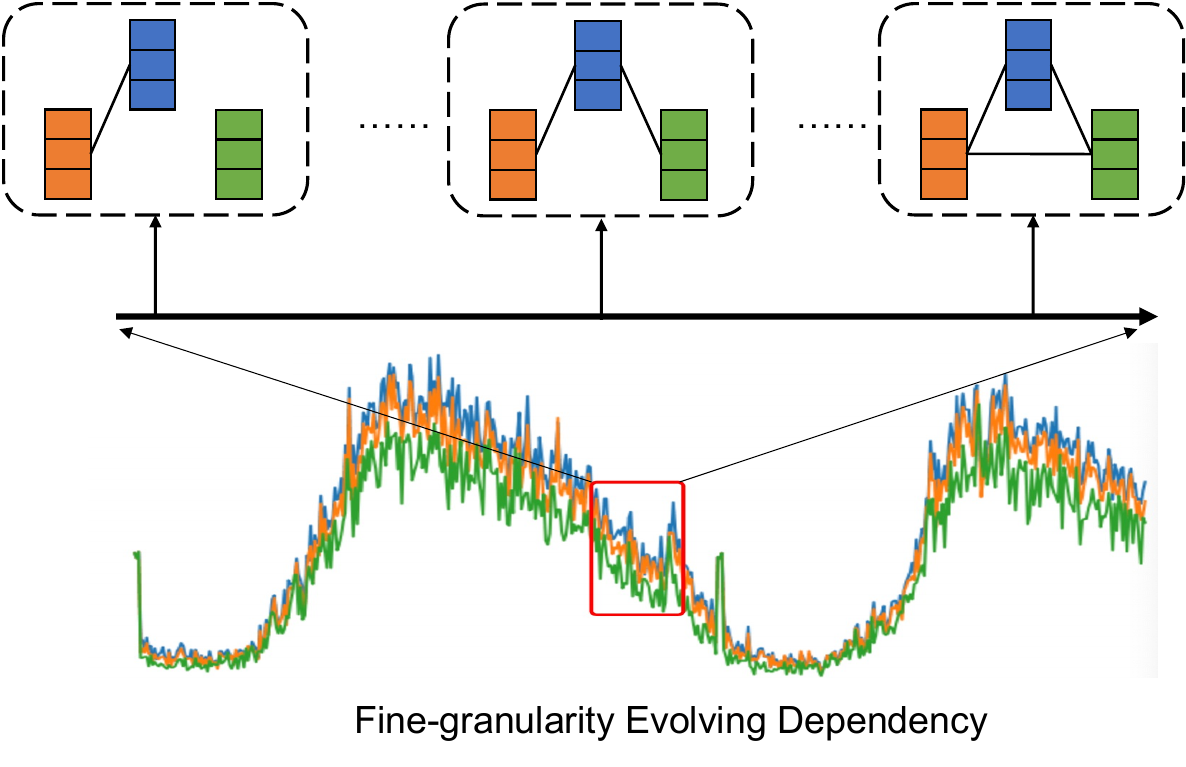}
\includegraphics[scale=0.4]{ 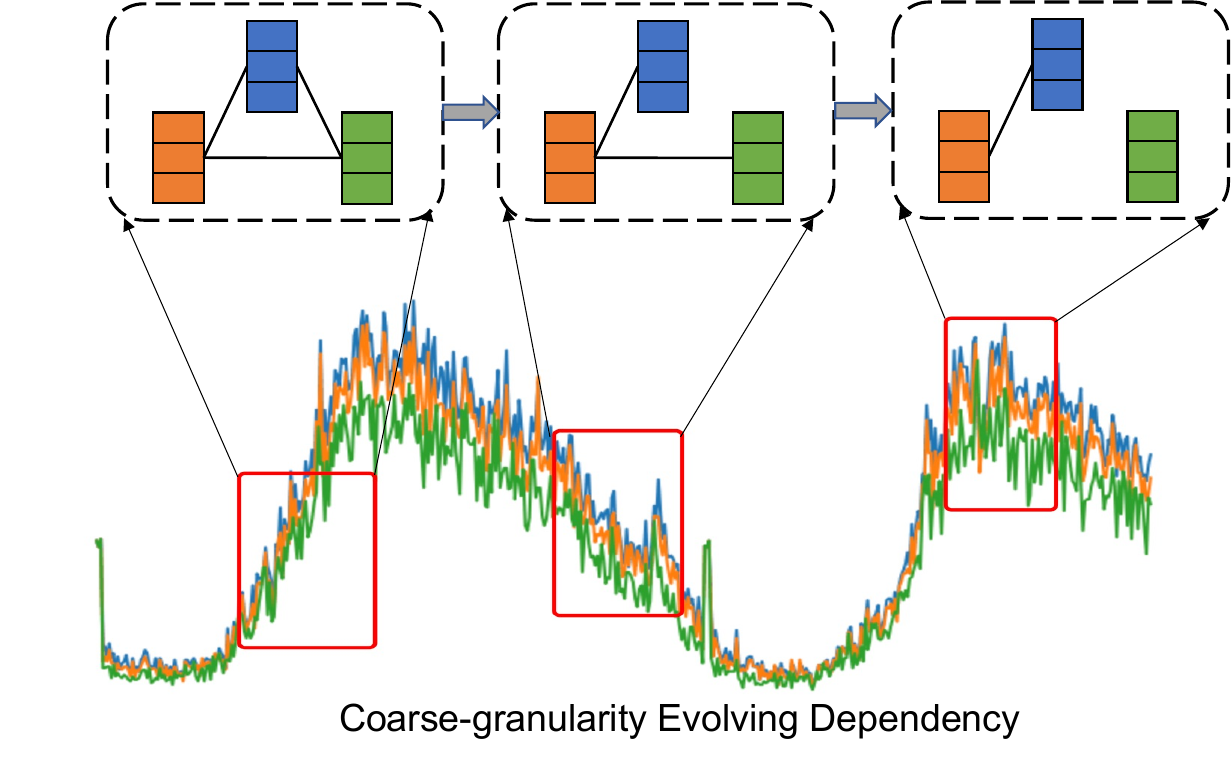}\\
\caption{The possible dynamic interactions of variables in TFP given different scales of observation. The red box means the sliding horizontal window, and vectors with different colors indicate  representation vectors of different series learned by the model.}\label{f1}
\end{figure}

Nevertheless, only a static adjacency matrix cannot present the dynamic temporal dependency in the road map. Recently, some models \citep{yu2017spatio, bai2019stg2seq} were proposed to replace the learnable series embedding matrix with the dynamic temporal dependency graph generated by time series encoders. As shown in Figure 1, there are normally two types of dynamic dependency, and the dynamic graph network is usually employed in existing works to model evolving dependency. As for the type of dynamic graph network \citep{kazemi2020representation}, discrete-time dynamic graph (DTDG) was normally restricted by the weak representing capacity in such high-flexibility micro-level dependency (e.g., fine-granularity dynamic deviation in each road) among traffic nodes. Thus, most of the graph-based methods in TFP preferred continuous-time dynamic graph (CTDG). However, because of the characteristics of GNN, the full graph is required to be stored and input into the model, and extra GNN-based processes have to be initialized in the TFP task. As a result, the computational time and space might be unaffordable if there are plenty of road nodes or long lengths of each flow series in the dataset. There were other works \citep{yu2022regularized} that extracted alternative information (e.g., shape-similarity and semantic-similarity among series) from implicitly similar nodes for enhancing the prediction. However, we argue that such kind of explicit spatial information, which reflects the temporal dependency in the spatial road map, was usually ignored in existing works.

In addition to graph-based methods, recent sequence-based works \citep{bai2018empirical, huang2019dsanet, lan2022dstagnn, STWA} mostly focused on capturing macro-temporal dependency (e.g., homomorphic wide-dynamic congestion in coarse-grained rush hour). In contrast, micro-level dependency lacks attention. Specifically, they extracted temporal dependencies based on the overall embedding of the whole series \citep{bai2018empirical, huang2019dsanet} or each timeslot \citep{lan2022dstagnn, STWA} after the encoding process, but ignored the deviation between nodes at each timestamp. One of the typical examples of hybrid temporal dependency is traffic congestion. For instance, if an accident occurs on the feeder road at 7:00 am, the resulting congestion can propagate and affect the highway's traffic flow well into the late morning. However, for the micro-temporal dependency, suppose there is a road segment near a busy intersection with traffic lights, the traffic flow in this segment can vary significantly at a granular level, such as minute-by-minute changes due to the traffic light cycles. Ignoring these micro-temporal dependencies makes the prediction cannot account for such cascading effects, decreasing the accuracy of traffic forecasts over extended periods.

Hence, to tackle the above issues in previous works, in this work, we proposed \textbf{S}patial-\textbf{T}emporal \textbf{A}ware \textbf{H}ybrid \textbf{G}raph Network (STAHGNet) to make predictions on traffic flow. Within this model, hybrid-granularity Spatial-temporal (HST) dependency is introduced and modeled. In other words, HST dependency is a type of heterogeneous Spatial-temporal correlation, and hybrid-granularity indicates both macro and micro-level dependencies are included. In particular, to provide macro-level information over the whole series and speed up the convergence of the model, feature engineering is used before the training of STAHGNet. Moreover, to capture the HST dependency, STAHGNet is designed as a recurrent model. The fine-grained temporal correlation is modeled at each timestamp and transmitted to the following modeling processes. The static spatial information is represented by both the inputted spatial traffic context and neighbor nodes. Specifically, the fine-grained temporal representation of each node is aggregated from spatial-context neighbor nodes by a heterogenous attention mechanism on the graph. Further, a random-sampling aggregation strategy based Hybrid Graph Attention mechanism (HGAT) is utilized in STAHGNet to resolve the expensive computational cost brought by GNN. Lastly, coarse-grained temporal information is modeled by our proposed Coarse-grained Temporal Graph (CTG) generator and further used to augment the final representation for prediction. Consequently, we conclude the contributions of this work as follows:
\begin{itemize}
\item To resolve the shortages and insufficiency of previous works in modeling HST dependency among traffic flows, STAHGNet is proposed to improve the GNN-based TFP methods by extracting evolving temporal dependency in hybrid-grained time scale and aggregating with static spatial features.
\item Correspondingly, an HGAT component is particularly designed to comprehensively aggregates static spatial information and continuously fine-grained temporal dependency between nodes based on a heterogenous attention mechanism. Besides, the random-sample strategy ensures this model has competitive efficiency.
\item An novel CTG component is implemented to generate the implicit auxiliary graph based on global representations of all training nodes. And to maintain the information from the temporal dimension, the attenuation coefficient replaces traditional operations (e.g., pooling) to aggregate representations effectively.
\item Extensive comparison experiments on four public datasets illustrated that our framework achieves superior results in both single and multi-step prediction compared to all baseline methods. And the necessity and effect of each crucial component in STAHGNet are validated by several ablation tests. 
\end{itemize} 

\section{Related Work}\label{sec2}

\subsection{Sequence-based model}\label{subsec1}
\noindent Conventional statistical models such as ARIMA \citep{3} and Vector Auto-Regression(VAR) \citep{zivot2006vector} have been used for TFP task, in which only intra-sequence relationships are considered, but not inter-sequence relationships. To make better prediction performance, some machine learning methods (e.g., SVM \citep{10}, XGBoost \citep{dong2018short}) were further exploited to model non-linear correlations within series. In recent research, deep learning methods gained a gradually elevated role in TFP. In general, to extract Spatio-temporal relations, there are three types of deep learning models that are commonly used: RNN-based \citep{gu2018recent}, CNN-based \citep{schuster1997bidirectional}, and Transformer-based \citep{vaswani2017attention}. The deep learning methods outperform traditional methods for TFP task, which involves time series prediction.

For RNN-based methods, except for basic RNN, LSTM \citep{hochreiter1997long} and GRU \citep{chung2014empirical} are also widely used for sequential dependency. FCLSTM \citep{sutskever2014sequence} is a typical model using LSTM to finish the whole modeling process. ConvLSTM \citep{shi2015convolutional} is a variant model of FCLSTM that extends of fully-connected LSTM, which converts the state-to-state calculations in LSTM to convolution, effectively solving the redundancy problem of LSTM in predicting Spatio-temporal data. On the other hand, CNN-based approaches were used to model the traffic map as an image consisting of a discontinuous grid where each grid contains spatial traffic features. TCN \citep{bai2018empirical} stacked several casual convolutional layers with exponentially enlarged dilation factors. DeepST \citep{zhang2016dnn} divides the time series into three subseries and convolves them separately. ST-ResNet \citep{zhang2017deep} replaces the convolution in DeepST with the residual convolution, which demonstrates the effectiveness of CNN for modeling Spatio-temporal data. Further, to enhance the fine-granularity dependency extraction from road maps, the attention mechanism has been extensively used. For example, DSANet \citep{huang2019dsanet} leveraged the CNN for the prediction and self-attention mechanism for spatial correlation modeling. STDN \citep{yao2019revisiting} uses local CNN and LSTM to handle the Spatio-temporal dependence of traffic flow data, introduces a flow-gate mechanism to learn dynamic similarity between locations, and designs a periodically shifted attention mechanism to handle long-term periodic temporal shifts. Among them, T-GCN \citep{zhao2019t} combined GRU and GCN to model fine-grained features at each time step, which is close to our proposed STAHGNet cell in Figure 3. Recently, STWave \cite{fang2023spatio} was proposed to take advantage of both the attention mechanism and convolution layer to learn the long-term trend. Nevertheless, it ignored global temporal and static spatial information, and whole graph modeling leads to more computational expense. 

Besides, Transformer-based methods show great power in sequence modeling. Informer \citep{zhou2021informer} extended the self-attention mechanism and took KL-divergence \citep{goldberger2003efficient} as the criterion to query dominant information in traffic flows. Local-sensitive hashing (LSH) was introduced by Reformer \citep{kitaev2020reformer} to approximate attention by allocating similar queries. However, although plenty of Transformer-based methods \citep{STWA,pmlr-v162-zhou22g, jin2023trafformer} have shown superior performance compared to CNN or RNN-based models, they normally suffer from quadratic memory and runtime overhead.

\subsection{Graph-based model}\label{subsec2}
\noindent Recently, many studies have attempted to use graph neural networks to model the correlation of spatial-temporal sequences in TFP. Graph convolutional network (GCN) \citep{zhao2019t} enables extracting high-level features of target road nodes by aggregating information from neighbor nodes. Specifically, categorized by the graph type, there are two types of methods in general. Spectral-type GCN \citep{bruna2013spectral} extended graph conventional convolution operation by Laplacian spectrum in the spectral domain. However, it suffers from expensive computational costs due to the calculation of all the eigenvalues of the Laplacian matrix. Thus, ChebNet \citep{defferrard2016convolutional} was
proposed to approximate the graph convolution by the Chebyshev-polynomial expansion of the eigenvalue diagonal matrix, and RGSL \citep{yu2022regularized} was proposed to model explicit spatial relation graph and extract implicit temporal graph simultaneously by a Laplacian matrix mixed-up module. Concentrating on TFP, DSTAGNN \citep{lan2022dstagnn} optimized the multi-head attention mechanism to capture dynamic spatial relevance and replace pre-defined static spatial graphs with learnable similarity graphs. However, extra computational cost brought by the implicit graph is inevitable and the built implicit graph is hard to interpret. 

Another type of GCN is based on spatial-type graphs. In \citep{micheli2009neural}, the neighborhood information was directly summarized directly. The Spatial-type GCN can measure not only spatial relationships but also dynamic temporal correlations. STGCN \citep{yu2017spatio} is a classical implementation of this type of GCN for temporal dependency, while its temporal GCN block cannot handle changes in graphs with the time flows. To tackle this problem in STG2Seq \citep{bai2019stg2seq}, a multiple-gated GCN was used as an encoder, and attention mechanisms were used as the decoder. Besides, STSGCN \citep{song2020spatial} was also proposed to compensate for the drawbacks in STGCN by introducing localized spatial-temporal information. But the graph encoders in both in STG2Seq and STSGCN remain GCN. To overcome the shortages of GCN in dynamic relationship modeling, Graph attention network (GAT) \citep{velickovic2017graph} used attention mechanisms to adjust the aggregation weights between nodes. Plenty of previous works deployed GAT as the dynamic dependency encoder \citep{guo2020short, zhang2019spatial}. MS-GAT \citep{huang2022learning} modeled spatial-temporal dependency and channel relation in traffic flow by a coupling GAT framework. There were also some works \cite{li2021spatial, fang2023spatio} harnessing the structural and semantic background knowledge inherent in both the traffic road network and historical traffic data using a spatiotemporal fusion graph. Nevertheless, many GCN-based methods fail to consider the fact that the correlations between sensors on the road network are dynamic and continuously evolving over time. Additionally, the notable memory and time consumption brought by the calculation among the full graph are always concerns in recent models. 

\begin{table}[h]
\caption{A summary of symbols and descriptions}\label{t1}
\scriptsize
\begin{tabular}{c|l} 
Symbol & Description\\
\hline
$\mathcal{G}$ & The traffic road network. \\
$\mathcal{V}$ & The road nodes set. \\
$\mathcal{E}$ & The edges set. \\
$\mathcal{F}$ & The expected trained mapping function to make predictions. \\
$N$ & The number of nodes in the $\mathcal{G}$. \\
$L$ & The length of the whole traffic flow. \\
$B$ & The size of a mini-batch. \\
$H$ & The hop number of sampled neighbor nodes. \\
$M$ & The number of subseries after the feature engineering. \\
$K$ & The number of sampled neighbor nodes. \\ 
$X$ & The finalized input series of each road node. \\ 
$G$ & The input series of graph-based neighbor nodes. \\ 
$D$ & The dimension of hidden vector. \\ 
$w$ & The size of the sliding historical window. \\
$A_s$ & The spatial weight adjacency matrix. \\
$A_t$ & The global temporal sparse adjacency matrix. \\
$\textbf{W}$ & The learnable parameter of the neural layer. \\ 
$\textbf{b}$ & The bias of the neural layer. \\ 
$x$ & The original series of the road node. \\ 
$x'$ & The normalized series of the road node. \\ 
$x''$ & The trend series of the road node. \\ 
$X$ & The integrated input series of the road node. \\ 
$y$ & The ground truth of each training series. \\ 
$\hat{y}$ & The output of STAHGNet. \\ 
$\textbf{c}$ & The state vector in the STAHGNet. \\
$\textbf{h}$ & The output temporal representation vector. \\
$\textbf{u}$ & The representation vectors of neighbor nodes. \\
$\textbf{r}$ & The output spatial-temporal representation vector by the last STAHGNet cell. \\
$\textbf{r}$ & The final spatial-temporal representation vector for prediction. \\
$\gamma$ & The attenuation coefficient. \\
$\alpha$ & The attention weight learned by attention mechanism. \\
$\mathcal{L}$ & The loss function used in this work. \\
\end{tabular}
\end{table}

\begin{table}[h]
\caption{A summary of acronyms in this paper}\label{acronym}
\scriptsize
\begin{tabular}{c|l} 
Acronym & Description\\
\hline
TFP & Traffic flow prediction. \\
ITS & Intelligent transportation system. \\
GNN & Graph neural network. \\
GCN & Graph convolutional network. \\
GAT&Graph attention network.\\
CNN & Convolutional neural network. \\
RNN & Recurrent neural network. \\
DTDG & Discrete-time dynamic graph. \\
CTDG & Continuous-time dynamic graph. \\
STAHGNet & Spatial-temporal aware hybrid graph network. \\
HST  & Hybrid-granularity Spatial-temporal. \\
HGAT &  Hybrid graph attention mechanism. \\
CTG & Coarse-grained temporal graph. \\ 
MLP & Multi-layered perceptron. \\ 
SD & Standard deviation. \\
\end{tabular}
\end{table}

\begin{figure*}
\begin{center}
\includegraphics[width=1\textwidth]{ 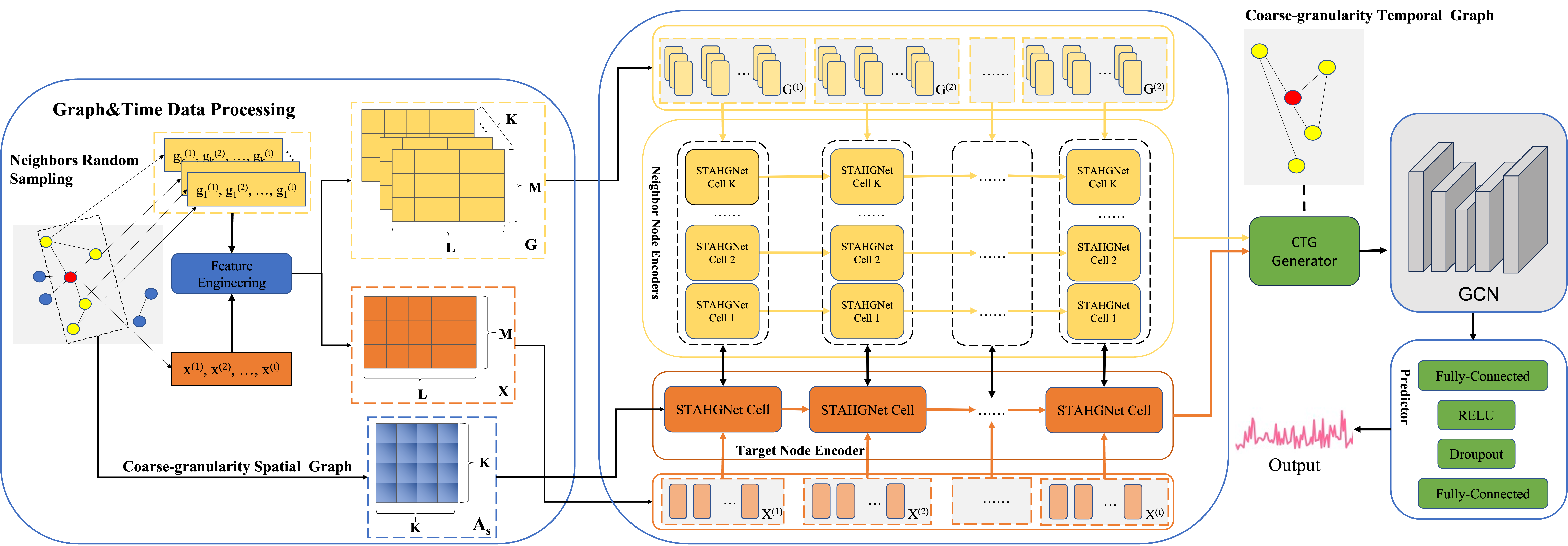}\\
\end{center}
\caption{The overall architecture of proposed STAHGNet. The single-step prediction is taken as the illustration example.}\label{f2}
\end{figure*}

\section{Methodology}\label{sec3}
\subsection{Problem Formulation}
\noindent The summary and descriptions of important notations in this work are shown in Table \ref{t1}, and the key acronyms in this work are summarized in Table \ref{acronym}. Firstly, we conceptualize the traffic road map as a graph $\mathcal{G} = (\mathcal{V}, \mathcal{E})$, where $\mathcal{V}$ consists of $N$ road nodes in this traffic network, and $\mathcal{E}$ is the set of connections between nodes. Considering the properties of traffic flow, for each road node $v$ in $\mathcal{V}$, the dynamic traffic volume denotes as $x = \{x^{(1)}, x^{(2)},..., x^{(t)}\}$, and $x^{(t)}$ is the value at the time slot $t$. Then, after our feature engineering, several additional series are supplemented. Thus, the inputted finalized series is $X \in \mathbb{R}^{L\times M}$, where $L$ indicates the length of the whole series and $M$ is the number of values at the $t$ timestamp. 

In this work, we focus on single-step prediction (i.e., taking one period of historical observations to make prediction on the state at the next time stamp after the period). Specifically, given input data $X$ and a sliding window with the size $w$, we divided it into $L-w$ subseries $X^{(t-w+1:t)}$, and take the original traffic volume $x^{(t+1)}$ as the ground truth $y$. A parameterized mapping function $\mathcal{F}$ is expected to learn in training for the prediction in this work.

\subsection{Model Architecture}
\noindent The overall architecture of the proposed STAHGNet is illustrated in Figure 2. Our model has three components: 1) Graph\&Time series processing; 2) Time series encoder; 3) Predictor. To be more specific, the Graph\&Time series processing aims to extract global features from the original flow series, and generate neighbor road nodes for the following feature-extracting processes. The Time series encoder component consists of two parts: several independent STAHGNets are initialized for each neighbor flow series; one STAHGNet is set for encoding the target road node. And the Predictor is expected to lean a map function from the feature vector $r^{(t+1)}$ to the prediction value $\hat{y}^{(t+1)}$. Detailed computational processes within STAHGNet are discussed in the following subsections.

\subsubsection{Automative Feature Engineering with Domain Knowledge} 
\noindent In the traffic scenario, some research \citep{lan2022dstagnn} validate that not all connections between road nodes reflect temporal dependency. Meanwhile, considering the high computational cost brought by GNNs  \citep{yu2022regularized,ye2022learning,lan2022dstagnn}, a random partial graph message passing is used in STAHGNet. Thus, in this component, we process the full graph structure into several subgraphs for each road node. Specifically, given one target road node $v$, we first collect all nodes with direct connections to $v$, and $K$ neighbor nodes are reserved. For the target node having less than $K$ neighbors, we compensate it for satisfying the demand of neighbor number $K$. Compared to previous GNN-based methods, the computational space for storing input data reduces from $\mathcal{O}{(B\times N\times w)}$ to $\mathcal{O}{(B\times K\times w)}$.

Then, inspired by \citep{elsayed2021we}, a feature engineering block is designed to enrich the input information provided for the supervised series predictor. An elaborately designed engineering can speed up the convergence of a sophisticated model and decipher the underlying inherent relationships within the dataset. There are plenty of feature engineering methods, including continuous series features, high-dimensional spatial transformation features, shapelet features, etc. Considering the strong capacity of deep learning methods to capture autocorrelation and periodical features, we focus on the continuous series feature method in this work.

Two types of continuous features are extracted from the univariate flow series of each node: 1) stability (i.e., normalized series $x'$); 2) trend (i.e., change rate series $x''$). Specifically, given a subseries $x^{(t-w+1:t)}$ after windowing the original univariate series, the normalized series is computed following (Eq.(1)), and the computation of change rate series refers to (Eq.(2)). Thus, through the data processing component, the finalized sequence is $X=\{(x^{(t)},x^{(t)\prime},x^{(t)\prime \prime})\}^L_{t=1}$, and the spatial-aware neighbor series is $G \in \mathbb{R}^{K\times L\times M}$.

\begin{equation}
\label{eq1}
x^{(i)\prime} = \frac{x^{(i)}-\operatorname{min}(x^{(t-w+1:t)})}{\operatorname{max}(x^{(t-w+1:t)})-\operatorname{min}(x^{(t-w+1:t)})},
\end{equation}
\begin{equation}
\label{eq2}
x^{(i)\prime \prime} = (x^{(i)}-x^{(i-1)})/x^{(i-1)}.
\end{equation}

\subsubsection{Spatio-Temporal Aware Hybrid Graph Network} 
\noindent The spatial-temporal dependency is critical to accurately measuring the correlation and distance between road nodes \citep{lan2022dstagnn}. Efforts of present state-of-the-art (SOTA) models concentrated on acquiring traffic temporal dependency given global spatial context and dynamic temporal graph structure, while there are still some challenges. Firstly, we argue that the continuously evolving temporal dependency graph learning is costly, and the implicit dependency without spatial structure between nodes is hard to interpret compared to the traffic road map. However, models considering only global or partially dynamic temporal dependency cannot capture fine-granularity correlation in traffic flows. In this section, we explain how to efficiently extract precise fine-granularity temporal dependency based on spatial graph structure.

\begin{figure}\centering
\includegraphics[scale=0.26]{ 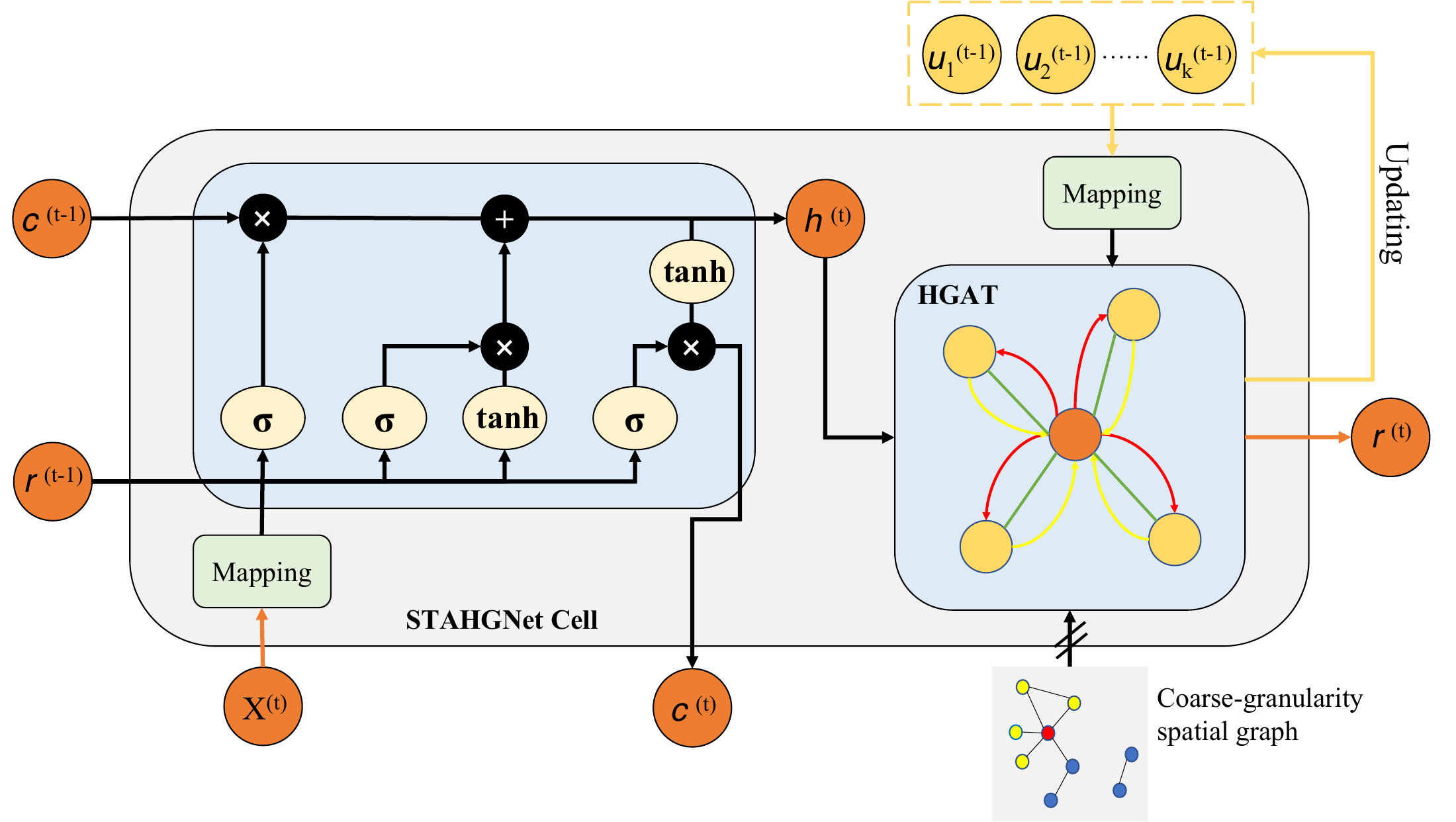} \\
\caption{Structure of STAHGNet cell. '$//$' on '$\rightarrow$' present stop-gradient. }\label{f3}
\end{figure}

Given flow series of target node $X \in \mathbb{R}^{L\times M}$ and its neighbor series $G \in \mathbb{R}^{K\times L\times M}$, $K+1$ independent STAHGNets are initialized for each series, and no parameter sharing between cells in each STAHGNet. The architecture of the STAHGNet cell is shown in  Figure 3. Specifically, at the $t$ time-stamp, the $X^{(t)}$ and $G^{(t)}$ are inputted into the first cell in their own encoder, respectively. For example, $(x^{(t)},x^{(t)}\prime,x^{(t)\prime \prime})$ is mapped into a high-dimension space to generate a representation vector $\mathbf{\hat{h}}^{(t)}$ with size $D$. A state vector $\textbf{c}^{(1)}\in \mathbb{R}^{1\times D}$ is randomly initialized at the first time slot and keeps updating through the sequence. The spatial-temporal aware representation $\textbf{r}^{(t-1)}$ at the last timestamp is also considered as one of the inputs at $t$ timestamp. Through a recurrent-based cell following Eq.(3), the state vector is updated to $\textbf{c}^{(t)}$ at each timestamp and prepared to enter the next cell. And the temporal information of this traffic flow at this time slot is represented by a vector $\textbf{h}^{(t)}\in \mathbb{R}^{1\times D}$.

\begin{equation}
\begin{aligned}
\label{eq3}
\mathbf{i}^{(t)} &= \sigma( \mathbf{W}_i^{T} [\mathbf{r}^{(t-1)} , \mathbf{\hat{h}}^{(t)} ] +\mathbf{b_i}), \\
\mathbf{f}^{(t)} &= \sigma(\mathbf{W}_f^{T}[\mathbf{r}^{(t-1)},\mathbf{\hat{h}}^{(t)}]+\mathbf{b_f}),\\
\mathbf{\hat{c}}^{(t)} &= \tanh(\mathbf{W}_c^{T}[\mathbf{r}^{(t-1)},\mathbf{\hat{h}}^{(t)}]+\mathbf{b_c}), \\
\mathbf{o}^{(t)} &= \sigma(\mathbf{W}_o^{T}[\mathbf{r}^{(t-1)},\mathbf{\hat{h}}^{(t)}]+\mathbf{b_o}), \\
\mathbf{c}^{(t)} &= \mathbf{f}^{(t)} \odot \mathbf{c}^{(t-1)} + \mathbf{i}^{(t)}\odot \mathbf{\hat{c}}^{(t)}, \\
\mathbf{h}^{(t)} &= \mathbf{o}^{(t)} \odot \tanh(\mathbf{c}^{(t)}).
\end{aligned}
\end{equation}

$\textbf{i}^{(t)}$ is input gate, $\textbf{f}^{(t)}$ is forget gate, and $\textbf{o}^{(t)}$ is output gate. $\textbf{W}_i$, $\textbf{W}_f$, $\textbf{W}_c$, $\textbf{W}_o \in \mathbb{R}^{D\times 2D}$  are the learnable parameter in each gate, $\textbf{b}_i$, $\textbf{b}_f$, $\textbf{b}_c$, $\textbf{b}_o$ is the bias, and $\odot$ indicates the element-wise product. After the temporal information extraction at both neighbor and target encoder sides, temporal representations of neighbor road nodes $\textbf{u}^{(t)}\in \mathbb{R}^{K\times D}$ and $\textbf{h}^{(t)}$ is put into the HGAT module, which is discussed in the next section.

\subsubsection{Hybrid Graph Attention Module}
\noindent GAT was proposed to overcome shortages of GCN: 1) GCN cannot conduct inductive tasks at the dynamic graph (i.e., it is hard to handle unseen nodes in test phase); 2) The topological structure of nodes is static, which means the temporal dependency cannot be captured. However, there are still some weaknesses in conventional GAT. Firstly, in the implementation of GAT, an extra adjacency matrix is required for each training sample, and all nodes are expected to be put into the model at the same time, which means a higher space and time cost. For example, if the HGAT in STAHGNet is replaced by GAT, the required input will change from $X\in \mathbb{R}^{L\times M}$ and $G\in \mathbb{R}^{K\times L\times M}$ to a larger sequence with a size of $L\times N\times M$. Besides, because of the property of the TFP task, the spatial information is normally consistent but significant for modeling correlation between nodes. Conventional GAT can only capture dynamic dependency but fails to maintain information from edges in the spatial graph.

To overcome the weaknesses of conventional GAT, we leverage a random-sampling information passing and heterogenous attention mechanism in the proposed HGAT to comprehend coarse-granularity spatial information and dynamic temporal dependency efficiently. Notes that the hybrid indicates both the hybrid-granularity attention mechanism and spatial-temporal awareness.

\textbf{Static Spatial Dependency} Firstly, at the first timestamp, the spatial graph $\mathcal{G}_s = (\mathcal{V}_s, \mathcal{E}_s)$ is prepared to be inputted into the HGAT block at the first STAHGNet cell. From the implementation perspective, a spatial weights matrix $A_s\in \mathbb{R}^{N\times N}$ presents the spatial weights, where each unit $\alpha_s$ in this matrix is computed based on the multiplicative inverse of spatial distance between nodes. Given target node $i$, after the data processing module, we mask units that are not selected as the neighbor nodes in $A_s$ at the column $i$. Thus, in our HGAT, the information aggregation of spatial dependency at each timestamp can refer to Eq.(4), where $\textbf{W}_s$ is the learnable weight in the spatial attention mechanism.

\begin{equation}
\label{eq4}
\mathbf{h}^{(t)}_s = \operatorname{RELU}(\mathbf{W}^{T}_s[\mathbf{h}^{(t)}, \sum^K_{i=1}\alpha_{si}\mathbf{u}^{(t)}_i]).
\end{equation}

\textbf{Fine-granularity Temporal Dependency} To capture continuous influence from neighbor nodes, the temporal attention score $\alpha_i^{(t)}$ between the target node and each neighbor is computed and updated at each timestamp. Given the temporal representation of the target node and neighbor nodes, the computational mechanism of information aggregation is shown in Eq.(5).

\begin{equation}
\begin{aligned}
\label{eq5}
e_i^{(t)} &= \mathbf{W}^{T}_q\mathbf{h}_s^{(t)} \times \mathbf{W}^{T}_k\mathbf{u}_i^{(t)}, \\
\alpha_i^{(t)} &= \frac{e_i^{(t)}}{\sum^K_{i=1}e_i^{(t)}}, \\
\mathbf{r}^{(t)} &= \operatorname{RELU}(\mathbf{W}^{T}_{fuse1}[\mathbf{h}^{(t)}, \sum^K_{i=1}\alpha_i^{(t)}\mathbf{W}^{T}_v\mathbf{u}_i^{(t)}]).
\end{aligned}
\end{equation}

In Eq.(5), $\textbf{W}_k$ is the learnable weight of the key mapping function, $\textbf{W}_q$ is the weight of the query function, and $\textbf{W}_v$ is the weight of the value function. After a linear transform by $\textbf{W}_k$, the representation matrix of neighbor nodes is queried by the query vector, which is transformed from the spatial representation of the target node. The attention score is used to weighted aggregate spatial information from neighbors. Then, through a fusion layer, we can obtain the updated spatial-temporal representation $\textbf{r}^{(t)}$ of the target node. After the information aggregation phase, the spatial-temporal information of the target node at the $t$ timestamp is passed to neighbor nodes as well. The detailed message-passing phase is implemented as Eq.(6), where $\textbf{W}_{fuse2}$ is the learnable weight of a new fusion layer.

\begin{equation}
\label{eq6}
\mathbf{u}_i^{(t)} = \operatorname{RELU}(\mathbf{W}^{T}_{fuse2}[\mathbf{h}^{(t)}, \mathbf{u}_i^{(t)}]).
\end{equation}

The message passing is operated on each neighbor node to update their representations. Then the updated spatial-temporal representation vectors of road nodes at the present timestamp are input to the STAHGNet cell for the next timestamp until the whole sequence is modeled.

\begin{figure}\centering
\includegraphics[scale=0.35]{ 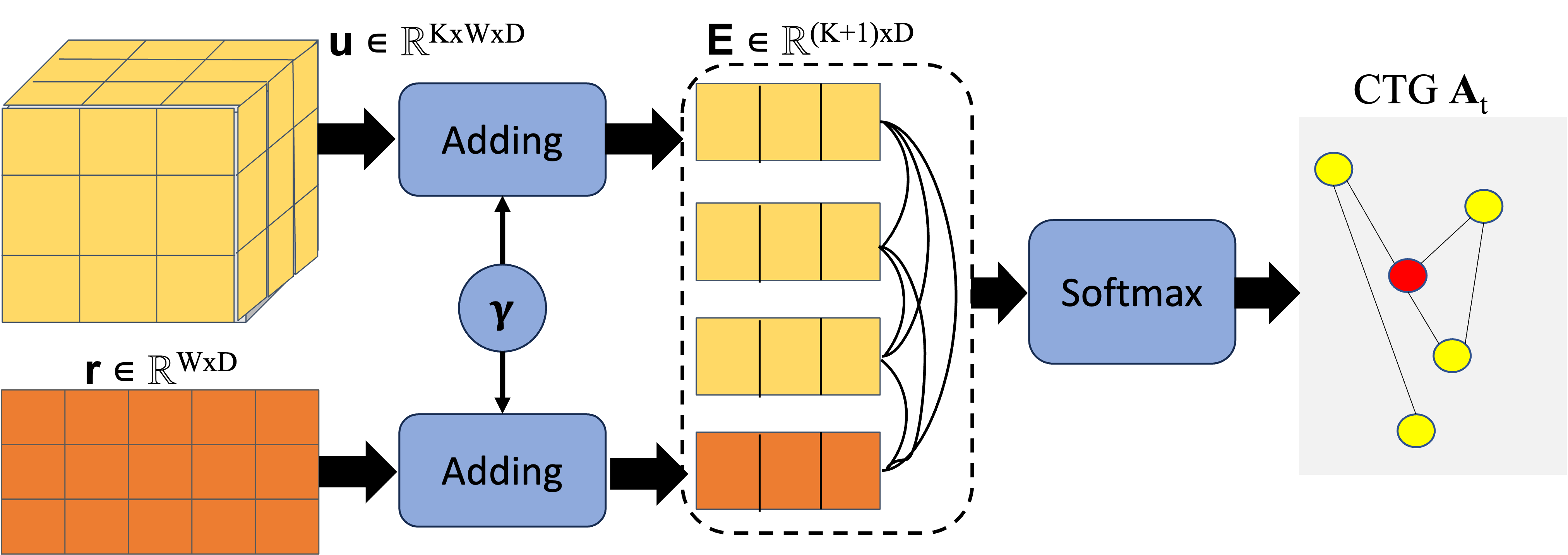} \\
\caption{Structure of CTG generator. }\label{f4}
\end{figure}

\textbf{Coarse-granularity Temporal Dependency}
Considering the inevitable global information loss during the recurrent modeling process, the final fine-grained Spatial-temporal representation $\textbf{r}$ is insufficient. Meanwhile, earlier research. Meanwhile, earlier research \citep{wu2019graph, bai2020adaptive} has shown that the fixed adjacency matrix may not accurately represent the underlying connections between nodes. Thus, we first design an implicit Coarse-granularity Temporal Graph (CTG) generator to structuralize evolving global temporal information. At the same time, because of the continuous property of the time series, a direct pooling operation or naive adding over representations might lose abundant information modeled in previous steps, or result in the redundancy in feature space. Concretely, as shown in Figure 4, given the representations at all time steps of both neighbor nodes $\textbf{u}$ and target node $\textbf{r}$, we first apply an adding operation with a computational attenuation coefficient $\gamma$ and aggregate information into the space $ \textbf{E} \in \mathbb{R}^{(K+1) \times D}$. With $\textbf{r}$ as an example, the process is formulated as follows:

\begin{equation}
\begin{aligned}
\label{eq7}
\gamma^{(t)} &= \frac{1}{w-t}, \\
\textbf{E} &= \operatorname{concat}[\sum_{t \in w}\gamma^{(t)}\textbf{r}^{(t)}, \sum_{t \in w}\gamma^{(t)}\textbf{u}^{(t)}_1], ..., \sum_{t \in w}\gamma^{(t)}\textbf{u}^{(t)}_N].
\end{aligned}
\end{equation}

Then, we apply a similarity matrix $\theta$, where the similarity between each node $theta_{ij}$ is measured by the inner product. Through a non-parameterized softmax function, a sparse adjacency matrix $A_t$ is computed as follows:

\begin{equation}
\label{eq8}
A_t = \sigma((\operatorname{log}(\theta_{ij}/(1-\theta_{ij}))
\end{equation}

where $\sigma$ is the activation function. Through CTG generator, we build an auxiliary implicit graph based on the global temporal representations. The $A_t$ is updated at each training iteration, which guarantees that the trainable graph remains statistically aligned with the evolving dependency matrix that is modifiable through training. Lastly, we select representations $r^{(t)}, u^{(t)}$ at the last time step $t$ with the $A_t$ into a graph convolutional network and further update the hidden spatial-temporal representation of the target node to $\textbf{r}^{(t)'}$. By the above operations, we suppose the new representation aggregates both hybrid-grained and spatial-temporal information consistently.

\subsubsection{Predictor and Loss Function}
\noindent The predictor is implemented by a Multi-Layered Perceptron (MLP) \citep{irie1988capabilities} to generate forecasting value $\hat{y}$ at $t+1$. In detail, it consists of two linear neural layers. To robustly transform the nonlinear dependencies to the future traffic flow, a non-linear activation function and a dropout operation \cite{srivastava2014dropout} are deployed between them as well. Through this MLP block, the output vector $\textbf{r}^{(t)'}$ of the target node encoder is transferred to the predict flow value $\hat{y}^{(t+1)}$ at $t+1$.  The smooth L1 loss is chosen as the loss function $\mathcal{L}$ of STAHGNet because of its robustness in the regression task. It is defined as follows:

\begin{equation}
\label{eq9}
\mathcal{L} = \begin{cases}
             0.5 \cdot (y-\hat{y})^{2}, & \text{if } |y-\hat{y}| < 0.5  \\  
             |y-\hat{y}| - 0.5, & \text{otherwise.}
             \end{cases}
\end{equation}

Let $\Theta$ denote all learnable parameters in STAHGNet, the optimization objective function is defined as Eq.(10). 

\begin{equation}
\label{eq10}
\mathop{\arg\min}_{\Theta}\mathcal{L}(X^{(t-w+1:t)}, G^{(t-w+1:t)}, A_s; x^{(t)}).
\end{equation}

\section{Expriments}\label{sec4}

In our experiments, we used three evaluation metrics to assess the performance of the proposed STAHGNet and multiple baselines on four public traffic flow datasets. In subsection 4.1, we provide details of the experiment setting. The results of the comparison experiment are presented in subsection 4.2. Subsection 4.3 discusses the sensitivity of our proposal to various parameters. We also analyze the effect of each key component of STAHGNet in subsection 4.4. Additionally, we conducted extra case studies to visualize the captured dynamic dependencies among road nodes in subsection 4.5 and to evaluate the overall computational costs of our model in subsection 4.6.

\subsection{Experimental Setup}
\subsubsection{Datasets}
To evaluate the performance of STAHGNet, we select four public traffic datasets: PeMSD3, PeMSD4, PeMSD7, and PeMSD8 \citep{song2020spatial}, which are both collected and issued by California Transportation Agencies Performance Measurement System\footnote{Data can be downloaded from https://pan.baidu.com/s/1ZPIiOM\_\_r1TRlmY4YGlolw with password p72z.}. The statistical summary of the datasets is shown in Table \ref{t2}. The flow data is aggregated into a 5-minute window, and the average value is presented. There is no personally identifying information (e.g., demographic of driver) in datasets, but the spatial distance between connected nodes is provided. The detailed statistical data description is shown in Table \ref{t2}.

\begin{table}[h]\centering
\caption{The Statistical Description of Datasets.}\label{t2}

\begin{tabular}{ccccc}
\hline
Dataset&Nodes&Edges&Timesteps&MissingRatio\\
\hline
PeMSD3& 358& 547& 26208& 0.672\% \\
PeMSD4& 307& 340& 16992& 3.182\% \\
PeMSD7& 883& 866& 28224& 0.452\% \\
PeMSD8& 170& 295& 17856& 0.696\% \\
\hline
\end{tabular}
\end{table}

\subsubsection{Baseline Methods}
Twelve classic or SOTA baseline methods are chosen to evaluate the effectiveness and superiority of the proposed STAHGNet. Within them, two categories can be specified: sequence models and graph models. 

In sequence models, they concentrate on extracting temporal correlations among series. \textbf{SVR} \citep{9}, a linear support vector machine is used for regression tasks. \textbf{ARIMA} \citep{3} is a classic statistics method modeling temporal dependencies within one series. \textbf{FCLSTM} \citep{sutskever2014sequence} uses LSTM as the series encoder for prediction. In \textbf{DSANet} \citep{huang2019dsanet}, the temporal and spatial correlations are captured by CNN and the self-attention mechanism, respectively. \textbf{TCN} \citep{bai2018empirical} learns local and global temporal relations hierarchically. \textbf{STFGNN} 
\cite{li2021spatial} designs a dynamic time warping-based temporal graph to extract spatial relationships that are functionally aware. In \textbf{ST-WA} \citep{STWA}, a window attention scheme is used to reduce complexity but still maintain performance. \textbf{MSSTRN} \cite{zhao2023multi}  introduces a data-driven method for generating a weighted adjacency matrix, effectively capturing real-time spatial dependencies that are not adequately captured by predefined matrices.

For graph models, \textbf{STG2Seq} \citep{bai2019stg2seq} designs a seq2seq architecture by a multi-gate GCN and attention mechanism. \textbf{STGCN} \citep{yu2017spatio} exploits spatial-temporal aware GCN in traffic forecasting. \textbf{STSGCN} \citep{song2020spatial} utilizes a GCN to capture the localized spatial-temporal correlations synchronously. \textbf{RGSL} \citep{yu2022regularized} fuse both implicit and explicit correlation graphs for forecasting. \textbf{MS-GAT} \citep{huang2022learning} exploits dynamic multi-aspect embeddings with TCN and attention mechanism to extract respective importance to prediction. \textbf{DSTAGNN} \citep{lan2022dstagnn} combined multi-head attention mechanism and multi-scale gated convolution to capture dynamic spatial-temporal dependency. \textbf{STWave} \cite{fang2023spatio} utilized a disentangle-fusion framework to mitigate the distribution shift in traffic data.

\subsubsection{Implementation Setting}
We adopt the Adam optimizer with a fixed learning rate of 0.0001 and the batch size is 64. The number of training epochs is set to 20, and the dimension of the hidden vector $D=64$. In addition, to avoid overfitting, the ratio of dropout units is 0.1. To obtain the best parameter combination, we use the grid search to tune parameters on the validation dataset. The candidate hyper-parameter value is: $K=[2,4,6,8]$, $w=[11, 16]$, and the sampled neighbor hop is $H=[1,2,3]$. All experiments are conducted on a 64-bit Linux server with GPU: NVIDIA GeForce GTX 3090. STAHGNet and all baselines are implemented based on Python 3.8 and PyTorch 1.7.0 \citep{paszke2019pytorch}. 

To fairly evaluate the performance of models, RMSE (Root Mean Square Error), MAE (Mean Absolute Error), and MAPE (Mean Absolute Percentage Error) are used to measure the difference between the ground truth and prediction value. We used the evaluation protocol outlined in \cite{song2020spatial} to split all to split all datasets into training (60\%), validation (20\%), and test (20\%) sets. We predicted the next hour’s data using one hour of historical data, meaning we used the past 12 continuous time steps (5 minutes for each step ) to predict the future 12 continuous time steps. Each experiment was repeated 5 times, and we used the paired t-test to ensure the performance of the best model was statistically better than that of the second-best model. In the tables, significant results (p-value$<$0.05) are bolded in black.

In the following sections, for those models that don't have available official codes, we implemented them ourselves; otherwise, their official codes are used. For fair comparisons, we extensively tuned their parameters to get the best performance.

\begin{sidewaystable}\centering
\renewcommand{\arraystretch}{1.5}
\caption{Overall Performance Comparison of Different Methods.}\label{t3}

\scriptsize
\begin{tabular}{c|ccc|ccc|ccc|ccc}
\toprule
& \multicolumn{3}{c|}{PeMSD3} & \multicolumn{3}{c|}{PeMSD4}& \multicolumn{3}{c|}{PeMSD7} & \multicolumn{3}{c}{PeMSD8} \\
Models &MAE            & RMSE           & MAPE(\%)       & MAE            & RMSE           & MAPE(\%) & MAE            & RMSE           & MAPE(\%)      & MAE            & RMSE  & MAPE(\%) \\
\hline
SVR      & 21.97          & 35.29          & 21.51          & 28.70          & 44.56          & 19.20    & 32.49          & 50.22          & 14.26         & 23.25          & 36.16 & 14.64    \\
                 ARIMA    & 35.41          & 47.59          & 33.78          & 33.73          & 48.80          & 24.18    & 38.17          & 59.27          & 19.46         & 31.09          & 44.32 & 22.73    \\
                 FCLSTM   & 21.33          & 35.11          & 23.33          & 26.77          & 40.65          & 18.23    & 29.98          & 45.94          & 13.20         & 23.09          & 35.17 & 14.99    \\
                 DSANet   & 21.29          & 34.55          & 23.21          & 22.79          & 35.77          & 16.03    & 31.36          & 49.11          & 14.43         & 17.14          & 26.96 & 11.32    \\
                 TCN      & 19.32          & 33.55          & 19.93          & 23.22          & 37.26          & 15.59    & 32.72          & 42.23          & 14.26         & 22.72          & 35.79 & 14.03    \\
                 STFGNN & 16.77 & 28.34 & 16.01  & 21.46 & 33.51 & 13.77  & 23.46 & 36.50 & 9.21  & 16.67 & 25.97 & 10.60  \\
                 ST-WA    & 15.71          & 27.53          & 15.90          & 19.56          & 31.02          & 12.73    & 21.02          & 35.01          & 9.10          & 15.95          & 25.11 & 10.05    \\
                 MSSTRN &15.10 & 26.54 & 14.07 & 19.03 & 30.95 & 12.76 & 20.34 & 34.29 & 8.53 & 15.43 & 24.77 & 9.97  \\
\hline
STG2Seq  & 19.03          & 29.83          & 21.55          & 25.20          & 38.48          & 18.77    & 32.77          & 47.16          & 20.16         & 20.17          & 30.71 & 17.32    \\
                 STGCN    & 17.49          & 30.12          & 17.15          & 22.70          & 35.55          & 14.59    & 25.38          & 38.78          & 11.08         & 18.02          & 27.83 & 11.40    \\
                 STSGCN   & 17.48          & 29.21          & 16.78          & 21.19          & 33.65          & 13.90    & 24.26          & 39.03          & 10.21         & 17.13          & 26.80 & 10.96    \\
                 MS-GAT   & 15.66          & 26.93          & 16.89          & 19.97          & 32.04          & 13.79    & 20.80          & 34.87          & 9.35          & 15.77          & 25.68 & 10.20    \\
                 RGSL     & 15.80          & 27.66          & 16.51          & 19.73          & 31.41          & 12.80    & 22.72          & 36.23          & 10.22         & 15.63          & 25.19 & 10.12    \\
                 DSTAGNN  & 16.17          & 28.12          & 15.68          & 19.49          & 30.86          & 12.83    & 21.60          & 34.89          & 9.61          & 16.07          & 24.96 & 10.24    \\
        STWave & 15.01 & 26.71 & 15.13  & 19.21 & 30.44 & 12.59  & 20.96 & 33.99 & 8.51  & 15.42 & 24.40 & 9.90  \\
\hline
STAHGNet & \textbf{14.82} & \textbf{25.01} & \textbf{13.37} & \textbf{18.92} & \textbf{29.62} & \textbf{12.35} & \textbf{20.02} & \textbf{33.68} & \textbf{8.02} & \textbf{15.29} & \textbf{24.15} & \textbf{9.37}          
    \\  
\toprule
\end{tabular}
\textnormal{Notes: In this table and the following tables, the statistically significant (p-value of the paired t-test $<$0.05) best results are bolded in \textbf{black}. The reported results of baselines are from our own testing but not cited from their reported results directly. }
\end{sidewaystable}

\begin{figure}\centering

\includegraphics[width=0.48\textwidth]{ 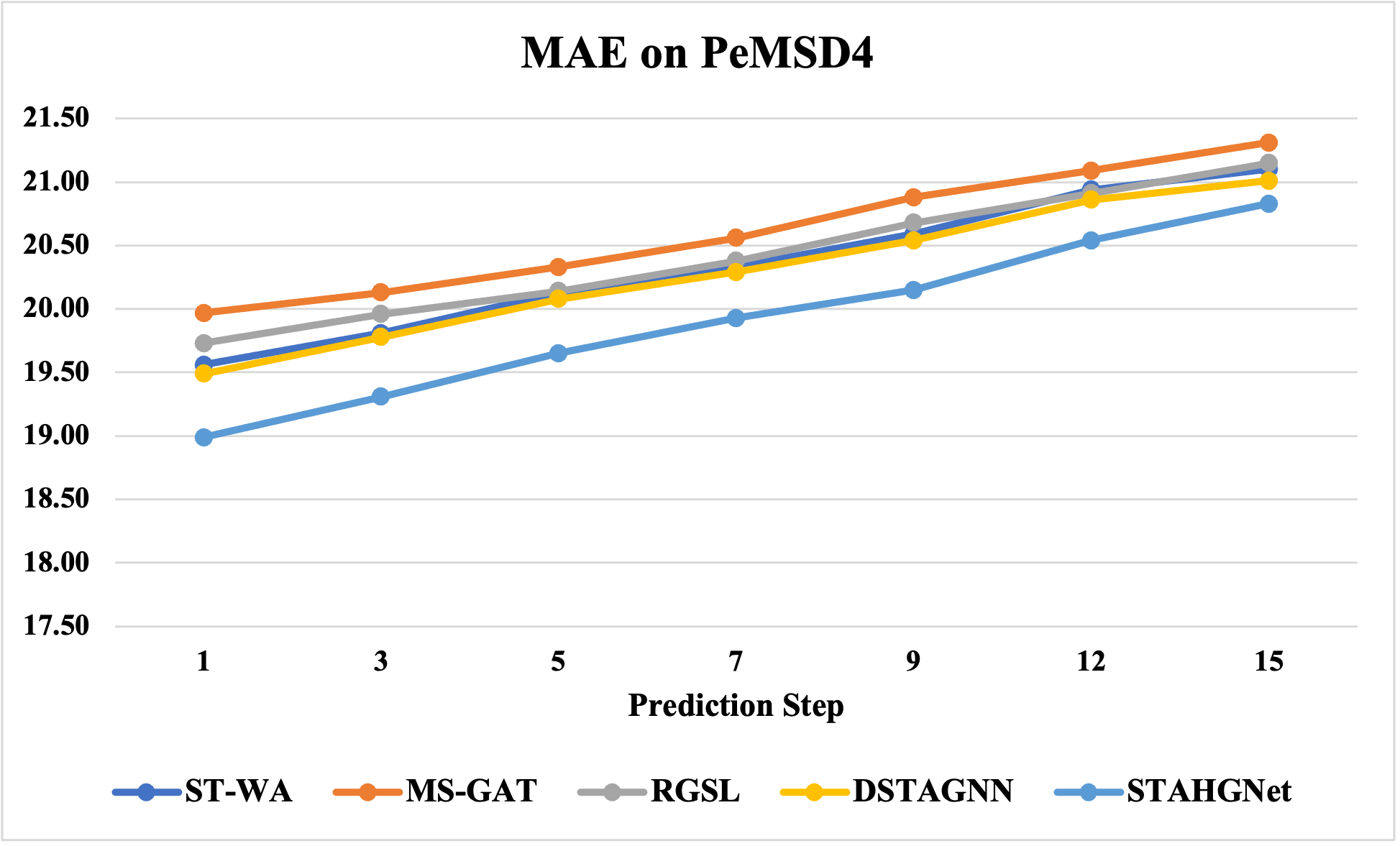}
\includegraphics[width=0.48\textwidth]{ 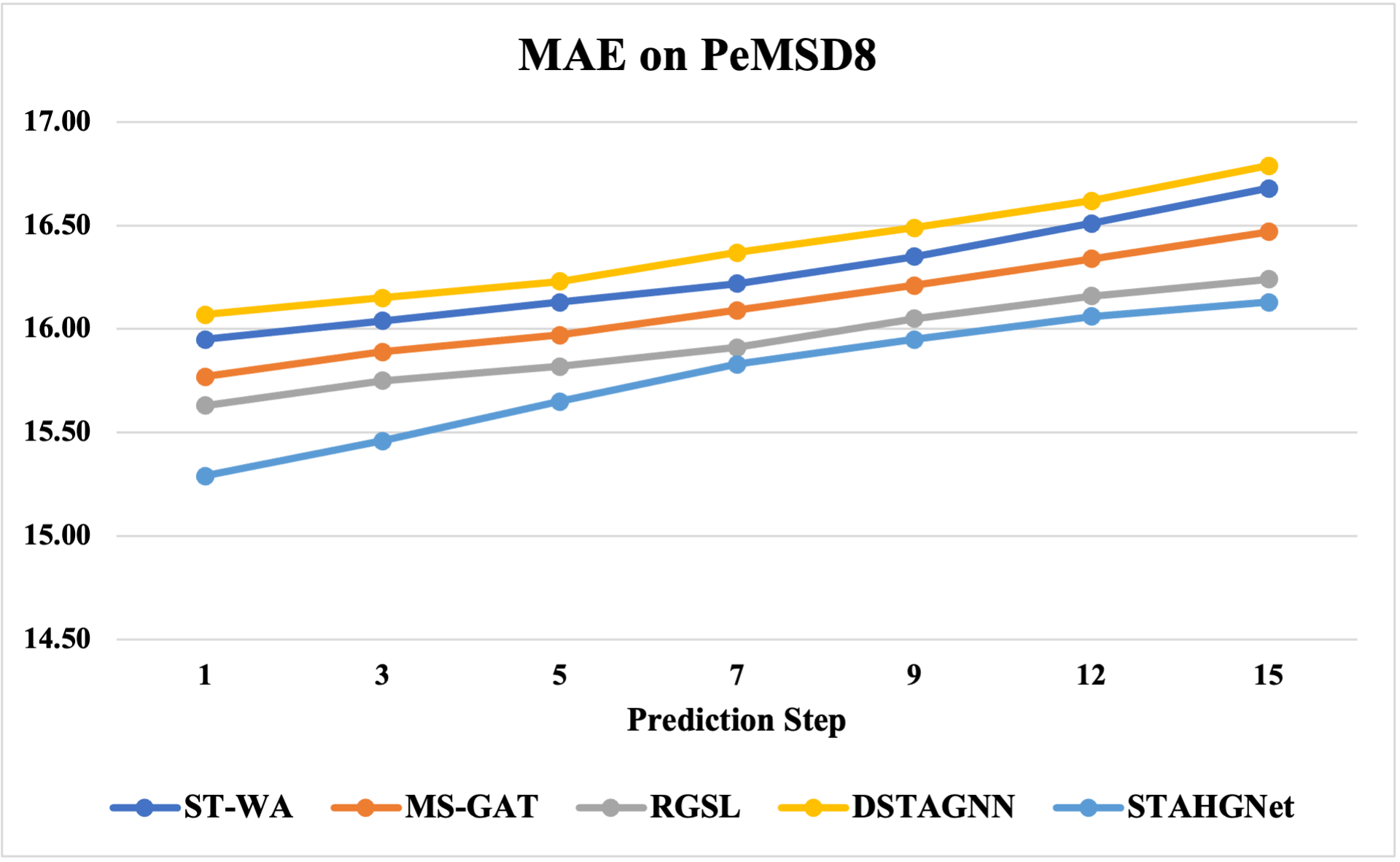}\\
\caption{Performance comparison on multi-step prediction.}\label{f5}
\end{figure}

\subsection{Comparison Results and Analysis}
\noindent Table \ref{t3} shows the results of the comparison experiment. Our STAHGNet consistently outperforms most baseline methods in all datasets, except for PeMSD4 and PeMSD8. In PeMSD4, the lowest MAPE is owned by RGSL, and the RMSE of STAHGNet is slightly higher (8.4\%) than DSTAGNN.

We can observe that sequence models have worse performance overall than graph models. It proves that the GNN structure is more suitable for modeling topological dependency. The traditional methods (i.e., SVR, ARIMA) always perform worse than deep-sequence methods (i.e., FCLSTM, DSANet, TCN), potentially due to the limited capacity to capture non-linear and complex relationships in traffic flows in traditional methods. By contrast, deep-sequence models are more capable of exploiting non-linear feature extraction. However, within deep-sequence models, the model with higher complexity does not always outperform others. For example, TCN has a lower MAE (19.32) in PeMSD3 than DSANet and FCLSTM, while FCLSTM (21.33) performs better than TCN. This indicates that the high complexity of the model cannot always lead to better prediction performance. 

Furthermore, earlier graph models (i.e., STG2Seq, STGCN, STSGCN) used a GCN module to capture spatial and temporal dependency, which may explain their relatively worse performance compared to STSGCN. Specifically, the STSGCN is able to leverage the localized dependency and boost the prediction performance. Finally, no significant gaps were observed among the models based on SOTA methods (i.e., DSTAGNN, ST-WA, MSSTRN, RGSL, MS-GAT, and STWave), and they all performed worse than the STAHGNet. This might be because although ST-WA, DSTAGNN, and MS-GAT leverage a multi-head attention structure to learn temporal dependency, the fine-granularity accumulated information is ignored. But our proposed STAHGNet exploits a recurrent network to dynamically capture spatial-temporal information and generate the representations of nodes at each timestamp. On the other, for RGSL or STWave, although a GCN models the fine-granularity deviation in each timestamp, the introduction of an extra implicit graph at each time step might bring unnecessary computational cost and features for TFP, and the static spatial graph is ignored in this work. By contrast, STAHGNet employs an HGAT to aggregate temporal and spatial correlations in fine granularity. Moreover, the usage of HGAT also makes STAHGNet maintain a competitive parameter size to ensure efficiency in training and testing, which are discussed in the following sections.

Moreover, referring to Figure 5, we conduct multi-step prediction over two datasets and compare performance with some SOTA baselines (i.e., DSTAGNN, ST-WA, RGSL, MS-GAT). We can first conclude that our proposal achieves the lowest prediction error in both two datasets. Additionally, we notice that, with the prediction step increase, the MAE lines of STAHGNet are gradually approaching baselines. Particularly when the prediction step increases to 9 in PeMSD4 and 7 in PeMSD8, the performance advantage of our model appears to be relatively reduced. Nevertheless, considering the periodic fluctuations or unforeseen anomalous values that may occur in the long-term prediction, past spatial-temporal dependency from historical sliding window data might not be adequate to present future long-term dependencies; thus the performance degradation of the model is to some extent unavoidable, which is also verified by the performance curves of baselines. Future work should focus on predicting long-term flows to address this issue.

\begin{table}
\caption{Performance Sensitivity to Hyper-Parameters.}\label{t4}
\centering
\scalebox{0.8}{\parbox{\linewidth}{%
    \begin{tabular}{c|c|ccc|ccc|ccc|ccc}
    \toprule
    \multirow{2}{*}{Metrics}  & \multirow{2}{*}{Param} & \multicolumn{3}{c|}{PeMSD3}            & \multicolumn{3}{c|}{PeMSD4}            & \multicolumn{3}{c|}{PeMSD7}            & \multicolumn{3}{c}{PeMSD8}             \\
                              &                        & 11    & 16    & SD                    & 11    & 16    & SD                    & 11    & 16    & SD                    & 11    & 16    & SD                     \\
    \hline
    \multirow{2}{*}{MAE}      & 1     & 15.06  & 15.02 & \multirow{2}{*}{0.20} & 20.00  & \textbf{18.17} & \multirow{2}{*}{0.60} & 20.34  & 20.70 & \multirow{2}{*}{0.65} & 16.19  & 16.06 & \multirow{2}{*}{0.40} \\
             & 2     & 15.23  & \textbf{14.82} &      & 20.54  & 19.48 &      & \textbf{20.02}  & 21.52 &      & \textbf{15.29}  & 15.84 &      \\
    \hline
    \multirow{2}{*}{RMSE}     & 1     & 25.74  & 25.24 & \multirow{2}{*}{0.33} & 30.15  & \textbf{29.62} & \multirow{2}{*}{0.24} & 33.87  & 34.68 & \multirow{2}{*}{0.44} & 24.92  & 24.74 & \multirow{2}{*}{0.41} \\
             & 2     & 24.58  & \textbf{25.01} &      & 30.03  & 30.07 &      & \textbf{33.68}  & 34.18 &      & \textbf{24.15}  & 24.63 &      \\
    \hline
    \multirow{2}{*}{MAPE(\%)} & 1     & 15.75  & 15.67 & \multirow{2}{*}{0.19} & 12.67  & \textbf{12.35} & \multirow{2}{*}{0.53} & 9.03  & 8.24  & \multirow{2}{*}{0.45} & 10.08  & 9.63  & \multirow{2}{*}{0.32} \\
             & 2     & 15.80  & \textbf{13.37} &      & 12.72  & 13.59 &      & \textbf{8.02}   & 8.21  &      & \textbf{9.37}   & 9.46   \\                     
            
    \toprule
    \end{tabular}
    }}
\textnormal{\\ Notes: For explored hyper-parameters, the horizontal axis indicates the size of historical window $w$, and the vertical axis indicates the hop $H$ of sampled neighbor nodes. The SD means the standard deviation of each group of parameter combinations. In this experiment, the number of sampled neighbors remains consistent with the number that obtains the best results in each dataset. }

\end{table}

\subsection{Parameter Sensitivity Test}

Table \ref{t4} shows the interaction effects of adjusting the hop number $H$ of sampled neighbors and historical window size $w$ on both four datasets. In Table \ref{t4}, the effect of hyper-parameter $H$ is presented at the vertical axis, and the effect of hyper-parameter $w$ can be seen at the horizontal axis. We notice that PeMSD3 has the overall lowest standard deviation (SD) in each metric, while PeMSD8 owns the highest SD. We suppose this is caused by the information volume of the dataset. The PeMSD8 has a relatively limited number of recorded road nodes and short flow length, the model can only learn an approximate $\Theta$ for further prediction phase. In addition, considering the difference between each dataset in Table \ref{t2}, the dataset with a long length of series (i.e., PeMSD3, PeMSD7) performs better when a larger historical window size. And for datasets with a high missing rate (i.e., PeMSD4), the prediction would be more accurate if the HGAT sampled neighbor nodes for two hops, while one hop is sufficient for the rest of the datasets in general. We argue that the missing value makes the signal-to-noise ratio low, and then a model with higher complexity is required to get better prediction performance. The information compensation provided by neighbor nodes can resolve this issue. The effect of sampled nodes is discussed in the next subsection.

\begin{sidewaystable}
\renewcommand{\arraystretch}{1.5}

\caption{Effect of HGAT, CTG generator, and random neighbor sampling Component.}\label{t5}
\centering
\scriptsize
\begin{tabular}{c|ccc|ccc|ccc|ccc}
\toprule
          & \multicolumn{3}{c|}{PeMSD3} & \multicolumn{3}{c|}{PeMSD4} & \multicolumn{3}{c|}{PeMSD7} & \multicolumn{3}{c}{PeMSD8}  \\
          & MAE & RMSE & MAPE(\%)      & MAE  & RMSE  & MAPE(\%)   & MAE   & RMSE  & MAPE(\%)   & MAE   & RMSE  & MAPE(\%)    \\
\hline
w/o $A_s$ & 16.38 &27.14	&16.85 &21.14	&31.81	&15.01	&22.14	&35.00	&11.86	&18.07	&28.44	&13.04       \\
w/o $A_t$ & 15.42 & 26.03 & 15.99 & 20.60 & 30.77 & 12.82 & 21.42 & 34.37 & 12.35 & 17.02 & 26.92 & 11.48 \\
$K$ = 0     & 16.41 & 27.26 & 16.92      & 21.37 & 32.05 & 15.17      & 22.75 & 35.12 & 12.59      & 18.73 & 29.20 & 13.35       \\
$K$ = 2     & 15.96 & 26.27 & 16.35      & 20.13 & 30.42 & 13.36      & 20.28 & 33.87 & 11.69      & 17.95 & 26.81 & 12.54       \\
$K$ = 4     & \textbf{14.82} & \textbf{25.01} & \textbf{13.37} & \textbf{18.92} & \textbf{29.62} & \textbf{12.35}      & 20.95 & 34.67 & 11.10      & \textbf{15.29} & \textbf{24.15} & \textbf{9.37}        \\
$K$ = 6     & 15.73 & 26.56 & 16.31      & 20.33 & 30.75 & 13.34      & \textbf{20.02} & \textbf{33.68} & \textbf{8.02}       & 16.30 & 26.76 & 11.12       \\
$K$ = 8     & 16.08 & 26.91 & 16.76      & 20.13 & 30.42 & 13.36      & 20.32 & 34.26 & 10.21      & 17.59 & 27.08 & 11.04       \\
$K$ = 10    & 15.61 & 26.40 & 16.25      & 20.33 & 30.39 & 13.37      & 20.44 & 34.19 & 9.69       & 17.56 & 26.97 & 11.81      
  \\   
\toprule
\end{tabular}
\textnormal{\\ Notes: For other hyper-parameters of STAHGNet on each dataset in this test, they are fixed, referring to Table \ref{t4} for fair comparisons. }
\end{sidewaystable}

\begin{figure*}\centering

\subfigure[Evolving dependency matrix of PeMSD3]{\centering
\includegraphics[scale=0.4]{ 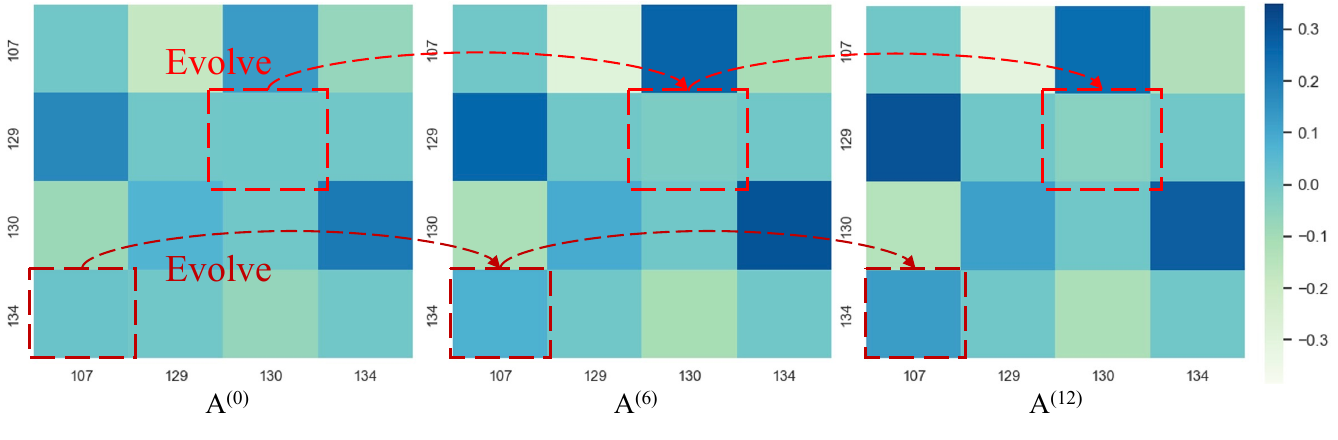} 
}
\subfigure[Evolving dependency matrix of PeMSD4]{\centering
\includegraphics[scale=0.4]{ 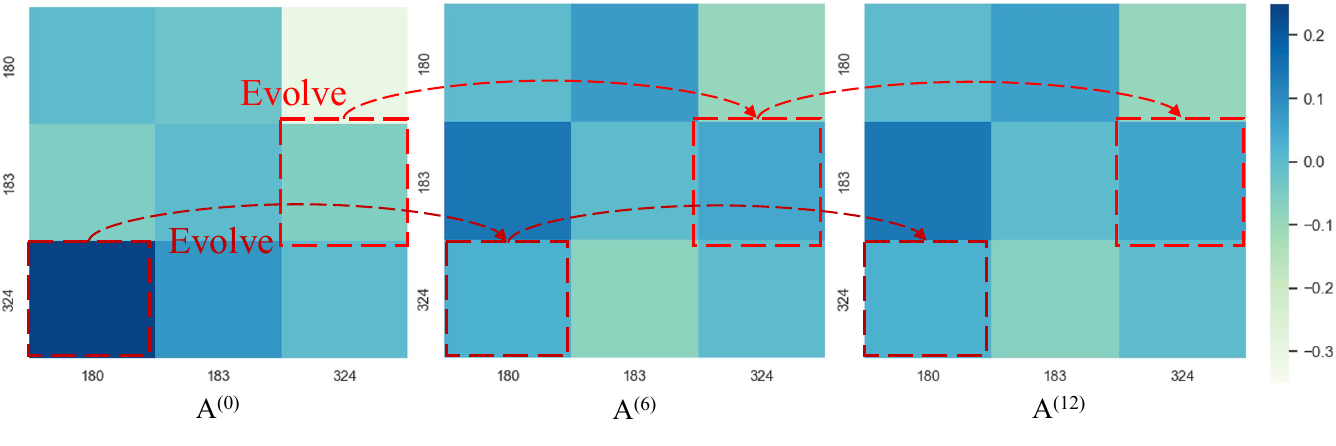} 
}
\subfigure[Historical curve of PeMSD3]{\centering
\includegraphics[scale=0.5]{ 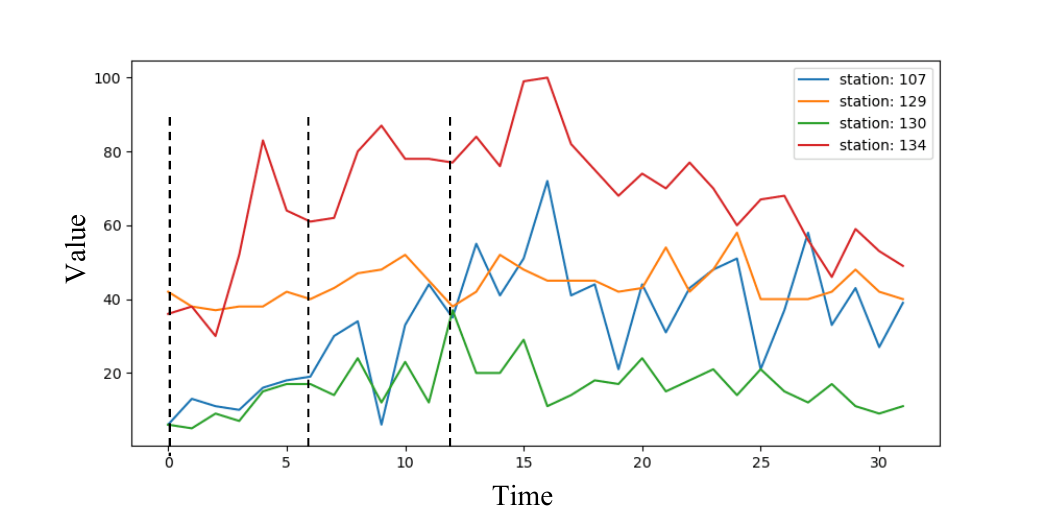} 
}
\subfigure[historical curve of PeMSD4]{\centering
\includegraphics[scale=0.5]{ 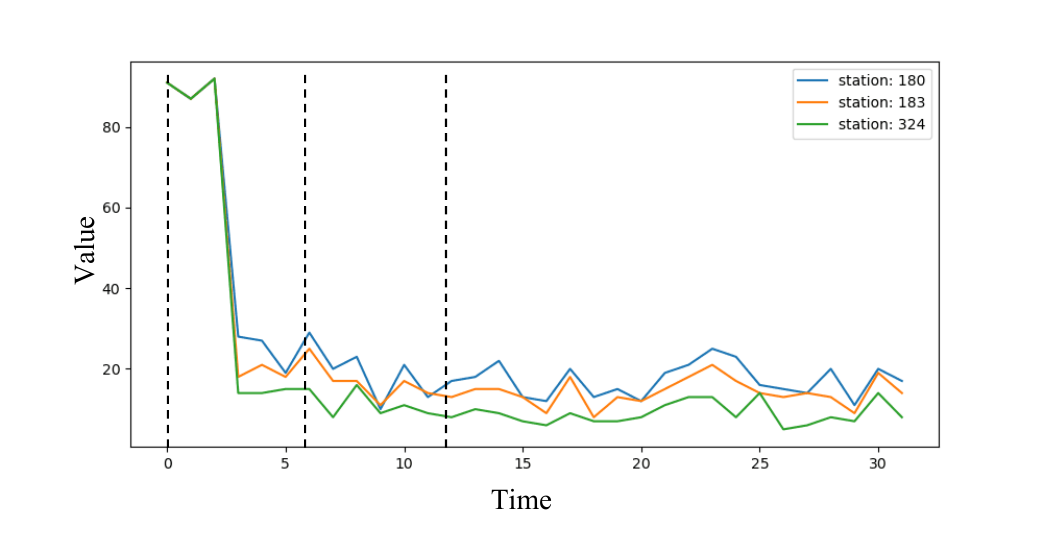} 
}

\caption{Each heatmap is computed at one timestamp. Each grid in (a) and (b) indicates the evolving correlation score captured by our HGAT component between two nodes.}\label{f6}
\end{figure*}

\subsection{Ablation Study}

The HGAT component aggregates two types of key information: hybrid-granularity temporal dependency and static spatial dependency. For temporal dependency extraction, we vary the sampled number of neighbor nodes $K$ from 0 to 10.  Sampled 0 nodes also means there is no temporal dependency explored. 

Seeing from Table \ref{t5}, as we increase the sampled number, the prediction performance does not become better. Particularly, the best result is normally given by $K=4$, except for PeMSD7. The reason is that the training of STAHGNet is harder with more neighbors involved. A proper sampled number is crucial to balance the performance and complexity. Nevertheless, abandoning the temporal dependency extraction still results in the worst performance. This proves the necessity of the fine-granularity temporal dependency in HGAT. Besides, after we remove the $A_s$, a decline in all metrics is observed on all datasets. This experiment confirms the effectiveness of combining fine-granularity temporal and static spatial dependency in HGAT. Meanwhile, if we eliminate the effect of $A_t$, we find that the performance of this variant is between the complete STAHGNet and the variant without $A_s$. In general, removing $A_s$ indicates the absence of spatial information. Therefore, we suppose that the introduction of the static spatial graph is necessary, and illustrates the importance of spatial features.

\subsection{Visualization of Evolving Dependency}

\noindent To investigate the effectiveness of our STAHGNet in modeling dynamic dependency, we select two datasets (i.e., PeMSD3 and PeMSD4) to conduct the case study. To present the dependency better, the smallest maximum connected graph in each dataset is chosen. As illustrated in Figure 6, Figure 6(a) and (b) present interaction matrix $A^{(0)}, A^{(6)}, A^{(12)}$ at three timestamps (i.e., 0, 6, and 12) for each dataset by heatmap, where the darker blue grid indicates higher weights. Figure 6(c)(d) displays the historical series curves. 

Firstly, as shown in Figure 6(a), the interaction score from stations 134 to 107 shows an increasing trend, and a lower interaction weight at timestamp 12 is observed. To look for validation, Figure 6(c) illustrates that when the values of nearby timestamps of nodes 107 and 134 get closer, the historical trends become more similar. Besides, for nodes 129 and 130, from time 0 to 6, there is no obvious trend change. But a notable mutation trend started around time 11, the value of station 129 declined, but the value of 130 grew higher. We also study the evolving correlations in PeMSD4. Similarly, for investigated interactions (nodes 324 and 180, nodes 183 and 324), the learned weight at timestamps 6 and 12 remains consistent, which can be proved by the similar value and trend gap around observed timestamps. Hence, the two phenomena above provide strong evidence to support the effectiveness of our learned evolving dependencies.

\subsection{Case Study of Computational Cost}

\begin{table}
\caption{The Comparison on Computational Cost.}\label{t6}
\centering
\begin{tabular}{c|ccccccc}
\toprule
                 & RGSL & ST-WA & MSSTRN& MS-GAT        & DSTAGNN & STWave& STAHGNet       \\
\hline
Space cost (MB)  & 7595 & 7515 &1319  & 5423          & 6635   &8587 & \textbf{1087}  \\
Time cost (it/s) & 1.43 & 2.00 & 2.09& \textbf{4.19} & 1.78  &1.11  & 2.15 \\
\toprule
\end{tabular}
\textnormal{\\ Notes: To quantify the computational cost in the training procedure. The space cost is measured by the Megabyte (MB) occupied in GPU memory, and the processed iteration number per second (it/s) is used to measure time cost.}
\end{table}

\noindent In this section, a specific case study on computational cost is conducted on the PeMSD4. As illustrated in Table \ref{t6}, to comprehensively verify the superior cost of our proposed model, both the space and time of SOTA methods are quantified for comparison. For fair comparisons, common hyper-parameters are kept consistent in all tested models. 

Observing Table \ref{t6}, STAHGNet owns the lowest GPU memory occupation among all methods. Specifically, MS-GAT shows a comparatively best performance in space cost among SOTA methods, while STAHGNet takes less than a quarter of the space compared to MS-GAT. Besides, MS-GAT has the highest processing speed (4.19 it/s). We argue that the low space cost of MS-GAT is because of the ignorance of implicit similarity graph construction. Meanwhile, the employed convolutional flow encoders in MS-GAT bring a higher processing speed, while it ignored fine-grained dependency modeling in its structural design, which makes MS-GAT perform less. By contrast, although STAHGNet captures both fine and coarse-grained dependency and remains the superior performance in TFP, our proposed model still outperforms other SOTA models. Even compared to MS-GAT, STAHGNet only has a speed lower by 48.69\%, which is significantly higher than the improvement in time-saving. Thus, we can conclude that our model maintains efficiency and performance well, and the effectiveness of feature engineering and random sampling for computational efficiency is validated as well.

\section{Conclusion}\label{sec13}
In this paper, we proposed a novel data-driven recurrent framework STAHGNet for TFP. In particular, an HGAT component and a CTG generator are designed and integrated into each STAHGNet cell to capture heterogeneously HST dependency. To balance the accuracy and efficiency, we deploy feature engineering before the training phase to speed up convergence, and a random sampling strategy is utilized in the Spatial-temporal graph construction to save space occupation and computational complexity. Extensive experiments on four public TFP datasets prove our proposal outperforms existing SOTA methods in both prediction performance and computational-friendly. Besides, the effectiveness and necessity of each key component in STAHGNet are verified by ablation tests and two case studies. The proposed TFP system can precisely predict future traffic, further achieve dynamic traffic guidance, and enhance the operational efficiency of the traffic system and the ability to actively prevent and control congestion. Modeling both long-term and fine-grained dependencies in TFP is fundamental for the development of intelligent transportation systems. The effectiveness of the random neighborhood road point sampling proposed in this paper also demonstrates the sparseness of the information contained in the traffic road network. In the future design of TFP systems, in addition to the need to model hybrid granularity spatio-temporal dependencies simultaneously, random neighbor road node sampling can be beneficial in building efficient and lightweight TFP systems.

However, given the precise dependency extraction at each timestamp, it is challenging to optimize the training time. Therefore, it is valuable to explore a more efficient way to implement such a complex recurrent model. Future work should focus on designing an alternative neural network structure that combines the benefits of recurrent networks in sequential modeling with the parallel modeling capabilities of transformer structures. Additionally, integrating the proposed STAHGNet or other TFP models into the intelligent transportation system, and deploying them in real environments efficiently and effectively will be an important future research direction of the TFP model.

\section*{Statements and Declarations}
\begin{itemize}
\item \textbf{Conflict of interest/Competing interests} The authors declare that they have no known competing financial interests or personal relationships that could have appeared to influence the work reported in this paper.
\item \textbf{Data availability} The four public traffic datasets (i.e., PeMSD3, PeMSD4, PeMSD7, and PeMSD8) are available at https://pan.baidu.com/s/1ZPIiOM\_\_r1TRlmY4YGlolw with password p72z. And all data supporting the findings of this study are available within the paper.
\item \textbf{Code availability} The code of this work will be released when it is prepared well.
\end{itemize}

\backmatter





\bibliography{sn-bibliography}


\begin{thebibliography}{52}
\ifx \bisbn   \undefined \def \bisbn  #1{ISBN #1}\fi
\ifx \binits  \undefined \def \binits#1{#1}\fi
\ifx \bauthor  \undefined \def \bauthor#1{#1}\fi
\ifx \batitle  \undefined \def \batitle#1{#1}\fi
\ifx \bjtitle  \undefined \def \bjtitle#1{#1}\fi
\ifx \bvolume  \undefined \def \bvolume#1{\textbf{#1}}\fi
\ifx \byear  \undefined \def \byear#1{#1}\fi
\ifx \bissue  \undefined \def \bissue#1{#1}\fi
\ifx \bfpage  \undefined \def \bfpage#1{#1}\fi
\ifx \blpage  \undefined \def \blpage #1{#1}\fi
\ifx \burl  \undefined \def \burl#1{\textsf{#1}}\fi
\ifx \doiurl  \undefined \def \doiurl#1{\url{https://doi.org/#1}}\fi
\ifx \betal  \undefined \def \betal{\textit{et al.}}\fi
\ifx \binstitute  \undefined \def \binstitute#1{#1}\fi
\ifx \binstitutionaled  \undefined \def \binstitutionaled#1{#1}\fi
\ifx \bctitle  \undefined \def \bctitle#1{#1}\fi
\ifx \beditor  \undefined \def \beditor#1{#1}\fi
\ifx \bpublisher  \undefined \def \bpublisher#1{#1}\fi
\ifx \bbtitle  \undefined \def \bbtitle#1{#1}\fi
\ifx \bedition  \undefined \def \bedition#1{#1}\fi
\ifx \bseriesno  \undefined \def \bseriesno#1{#1}\fi
\ifx \blocation  \undefined \def \blocation#1{#1}\fi
\ifx \bsertitle  \undefined \def \bsertitle#1{#1}\fi
\ifx \bsnm \undefined \def \bsnm#1{#1}\fi
\ifx \bsuffix \undefined \def \bsuffix#1{#1}\fi
\ifx \bparticle \undefined \def \bparticle#1{#1}\fi
\ifx \barticle \undefined \def \barticle#1{#1}\fi
\bibcommenthead
\ifx \bconfdate \undefined \def \bconfdate #1{#1}\fi
\ifx \botherref \undefined \def \botherref #1{#1}\fi
\ifx \url \undefined \def \url#1{\textsf{#1}}\fi
\ifx \bchapter \undefined \def \bchapter#1{#1}\fi
\ifx \bbook \undefined \def \bbook#1{#1}\fi
\ifx \bcomment \undefined \def \bcomment#1{#1}\fi
\ifx \oauthor \undefined \def \oauthor#1{#1}\fi
\ifx \citeauthoryear \undefined \def \citeauthoryear#1{#1}\fi
\ifx \endbibitem  \undefined \def \endbibitem {}\fi
\ifx \bconflocation  \undefined \def \bconflocation#1{#1}\fi
\ifx \arxivurl  \undefined \def \arxivurl#1{\textsf{#1}}\fi
\csname PreBibitemsHook\endcsname

\bibitem[\protect\citeauthoryear{Yang et~al.}{2021}]{yang2021unsupervised}
\begin{botherref}
\oauthor{\bsnm{Yang}, \binits{S.B.}},
\oauthor{\bsnm{Guo}, \binits{C.}},
\oauthor{\bsnm{Hu}, \binits{J.}},
\oauthor{\bsnm{Tang}, \binits{J.}},
\oauthor{\bsnm{Yang}, \binits{B.}}:
Unsupervised path representation learning with curriculum negative sampling.
arXiv preprint arXiv:2106.09373
(2021)
\end{botherref}
\endbibitem

\bibitem[\protect\citeauthoryear{Smith et~al.}{2002}]{1}
\begin{barticle}
\bauthor{\bsnm{Smith}, \binits{B.L.}},
\bauthor{\bsnm{Williams}, \binits{B.M.}},
\bauthor{\bsnm{Oswald}, \binits{R.K.}}:
\batitle{Comparison of parametric and nonparametric models for traffic flow forecasting}.
\bjtitle{Transportation research. Part C}
\bvolume{10C}(\bissue{4}),
\bfpage{303}--\blpage{321}
(\byear{2002})
\end{barticle}
\endbibitem

\bibitem[\protect\citeauthoryear{Okutani and Stephanedes}{1984}]{2}
\begin{botherref}
\oauthor{\bsnm{Okutani}, \binits{I.}},
\oauthor{\bsnm{Stephanedes}, \binits{Y.J.}}:
Dynamic prediction of traffic volume through kalman filtering
\textbf{18}(1),
1--11
(1984)
\end{botherref}
\endbibitem

\bibitem[\protect\citeauthoryear{Voort et~al.}{1996}]{3}
\begin{barticle}
\bauthor{\bsnm{Voort}, \binits{M.}},
\bauthor{\bsnm{Dougherty}, \binits{M.}},
\bauthor{\bsnm{Watson}, \binits{S.}}:
\batitle{Combining kohonen maps with arima time series models to forecast traffic flow}.
\bjtitle{Transportation Research Part C Emerging Technologies}
\bvolume{4}(\bissue{5}),
\bfpage{307}--\blpage{318}
(\byear{1996})
\end{barticle}
\endbibitem

\bibitem[\protect\citeauthoryear{Cai et~al.}{2016}]{6}
\begin{barticle}
\bauthor{\bsnm{Cai}, \binits{P.}},
\bauthor{\bsnm{Wang}, \binits{Y.}},
\bauthor{\bsnm{Lu}, \binits{G.}},
\bauthor{\bsnm{Chen}, \binits{P.}},
\bauthor{\bsnm{Ding}, \binits{C.}},
\bauthor{\bsnm{Sun}, \binits{J.}}:
\batitle{A spatiotemporal correlative k-nearest neighbor model for short-term traffic multistep forecasting}.
\bjtitle{Transportation Research Part C}
\bvolume{62},
\bfpage{21}--\blpage{34}
(\byear{2016})
\end{barticle}
\endbibitem

\bibitem[\protect\citeauthoryear{Hong et~al.}{2011}]{9}
\begin{botherref}
\oauthor{\bsnm{Hong}, \binits{W.C.}},
\oauthor{\bsnm{Dong}, \binits{Y.}},
\oauthor{\bsnm{Zheng}, \binits{F.}},
\oauthor{\bsnm{Lai}, \binits{C.Y.}}:
Forecasting urban traffic flow by svr with continuous aco.
Applied mathematical modelling
(3),
35
(2011)
\end{botherref}
\endbibitem

\bibitem[\protect\citeauthoryear{Gu et~al.}{2018}]{gu2018recent}
\begin{barticle}
\bauthor{\bsnm{Gu}, \binits{J.}},
\bauthor{\bsnm{Wang}, \binits{Z.}},
\bauthor{\bsnm{Kuen}, \binits{J.}},
\bauthor{\bsnm{Ma}, \binits{L.}},
\bauthor{\bsnm{Shahroudy}, \binits{A.}},
\bauthor{\bsnm{Shuai}, \binits{B.}},
\bauthor{\bsnm{Liu}, \binits{T.}},
\bauthor{\bsnm{Wang}, \binits{X.}},
\bauthor{\bsnm{Wang}, \binits{G.}},
\bauthor{\bsnm{Cai}, \binits{J.}}, \betal:
\batitle{Recent advances in convolutional neural networks}.
\bjtitle{Pattern recognition}
\bvolume{77},
\bfpage{354}--\blpage{377}
(\byear{2018})
\end{barticle}
\endbibitem

\bibitem[\protect\citeauthoryear{Schuster and Paliwal}{1997}]{schuster1997bidirectional}
\begin{barticle}
\bauthor{\bsnm{Schuster}, \binits{M.}},
\bauthor{\bsnm{Paliwal}, \binits{K.K.}}:
\batitle{Bidirectional recurrent neural networks}.
\bjtitle{IEEE transactions on Signal Processing}
\bvolume{45}(\bissue{11}),
\bfpage{2673}--\blpage{2681}
(\byear{1997})
\end{barticle}
\endbibitem

\bibitem[\protect\citeauthoryear{Vaswani et~al.}{2017}]{vaswani2017attention}
\begin{botherref}
\oauthor{\bsnm{Vaswani}, \binits{A.}},
\oauthor{\bsnm{Shazeer}, \binits{N.}},
\oauthor{\bsnm{Parmar}, \binits{N.}},
\oauthor{\bsnm{Uszkoreit}, \binits{J.}},
\oauthor{\bsnm{Jones}, \binits{L.}},
\oauthor{\bsnm{Gomez}, \binits{A.N.}},
\oauthor{\bsnm{Kaiser}, \binits{{\L}.}},
\oauthor{\bsnm{Polosukhin}, \binits{I.}}:
Attention is all you need.
Advances in neural information processing systems
\textbf{30}
(2017)
\end{botherref}
\endbibitem

\bibitem[\protect\citeauthoryear{Yu et~al.}{2017}]{yu2017spatio}
\begin{botherref}
\oauthor{\bsnm{Yu}, \binits{B.}},
\oauthor{\bsnm{Yin}, \binits{H.}},
\oauthor{\bsnm{Zhu}, \binits{Z.}}:
Spatio-temporal graph convolutional networks: A deep learning framework for traffic forecasting.
arXiv preprint arXiv:1709.04875
(2017)
\end{botherref}
\endbibitem

\bibitem[\protect\citeauthoryear{Zhao et~al.}{2019}]{zhao2019t}
\begin{barticle}
\bauthor{\bsnm{Zhao}, \binits{L.}},
\bauthor{\bsnm{Song}, \binits{Y.}},
\bauthor{\bsnm{Zhang}, \binits{C.}},
\bauthor{\bsnm{Liu}, \binits{Y.}},
\bauthor{\bsnm{Wang}, \binits{P.}},
\bauthor{\bsnm{Lin}, \binits{T.}},
\bauthor{\bsnm{Deng}, \binits{M.}},
\bauthor{\bsnm{Li}, \binits{H.}}:
\batitle{T-gcn: A temporal graph convolutional network for traffic prediction}.
\bjtitle{IEEE Transactions on Intelligent Transportation Systems}
\bvolume{21}(\bissue{9}),
\bfpage{3848}--\blpage{3858}
(\byear{2019})
\end{barticle}
\endbibitem

\bibitem[\protect\citeauthoryear{Scarselli et~al.}{2008}]{scarselli2008graph}
\begin{barticle}
\bauthor{\bsnm{Scarselli}, \binits{F.}},
\bauthor{\bsnm{Gori}, \binits{M.}},
\bauthor{\bsnm{Tsoi}, \binits{A.C.}},
\bauthor{\bsnm{Hagenbuchner}, \binits{M.}},
\bauthor{\bsnm{Monfardini}, \binits{G.}}:
\batitle{The graph neural network model}.
\bjtitle{IEEE transactions on neural networks}
\bvolume{20}(\bissue{1}),
\bfpage{61}--\blpage{80}
(\byear{2008})
\end{barticle}
\endbibitem

\bibitem[\protect\citeauthoryear{Bai et~al.}{2019}]{bai2019stg2seq}
\begin{botherref}
\oauthor{\bsnm{Bai}, \binits{L.}},
\oauthor{\bsnm{Yao}, \binits{L.}},
\oauthor{\bsnm{Kanhere}, \binits{S.}},
\oauthor{\bsnm{Wang}, \binits{X.}},
\oauthor{\bsnm{Sheng}, \binits{Q.}}, et al.:
Stg2seq: Spatial-temporal graph to sequence model for multi-step passenger demand forecasting.
arXiv preprint arXiv:1905.10069
(2019)
\end{botherref}
\endbibitem

\bibitem[\protect\citeauthoryear{Kazemi et~al.}{2020}]{kazemi2020representation}
\begin{barticle}
\bauthor{\bsnm{Kazemi}, \binits{S.M.}},
\bauthor{\bsnm{Goel}, \binits{R.}},
\bauthor{\bsnm{Jain}, \binits{K.}},
\bauthor{\bsnm{Kobyzev}, \binits{I.}},
\bauthor{\bsnm{Sethi}, \binits{A.}},
\bauthor{\bsnm{Forsyth}, \binits{P.}},
\bauthor{\bsnm{Poupart}, \binits{P.}}:
\batitle{Representation learning for dynamic graphs: A survey.}
\bjtitle{J. Mach. Learn. Res.}
\bvolume{21}(\bissue{70}),
\bfpage{1}--\blpage{73}
(\byear{2020})
\end{barticle}
\endbibitem

\bibitem[\protect\citeauthoryear{Yu et~al.}{2022}]{yu2022regularized}
\begin{botherref}
\oauthor{\bsnm{Yu}, \binits{H.}},
\oauthor{\bsnm{Li}, \binits{T.}},
\oauthor{\bsnm{Yu}, \binits{W.}},
\oauthor{\bsnm{Li}, \binits{J.}},
\oauthor{\bsnm{Huang}, \binits{Y.}},
\oauthor{\bsnm{Wang}, \binits{L.}},
\oauthor{\bsnm{Liu}, \binits{A.}}:
Regularized graph structure learning with semantic knowledge for multi-variates time-series forecasting.
arXiv preprint arXiv:2210.06126
(2022)
\end{botherref}
\endbibitem

\bibitem[\protect\citeauthoryear{Bai et~al.}{2018}]{bai2018empirical}
\begin{botherref}
\oauthor{\bsnm{Bai}, \binits{S.}},
\oauthor{\bsnm{Kolter}, \binits{J.Z.}},
\oauthor{\bsnm{Koltun}, \binits{V.}}:
An empirical evaluation of generic convolutional and recurrent networks for sequence modeling.
arXiv preprint arXiv:1803.01271
(2018)
\end{botherref}
\endbibitem

\bibitem[\protect\citeauthoryear{Huang et~al.}{2019}]{huang2019dsanet}
\begin{bchapter}
\bauthor{\bsnm{Huang}, \binits{S.}},
\bauthor{\bsnm{Wang}, \binits{D.}},
\bauthor{\bsnm{Wu}, \binits{X.}},
\bauthor{\bsnm{Tang}, \binits{A.}}:
\bctitle{Dsanet: Dual self-attention network for multivariate time series forecasting}.
In: \bbtitle{Proceedings of the 28th ACM International Conference on Information and Knowledge Management},
pp. \bfpage{2129}--\blpage{2132}
(\byear{2019})
\end{bchapter}
\endbibitem

\bibitem[\protect\citeauthoryear{Lan et~al.}{2022}]{lan2022dstagnn}
\begin{bchapter}
\bauthor{\bsnm{Lan}, \binits{S.}},
\bauthor{\bsnm{Ma}, \binits{Y.}},
\bauthor{\bsnm{Huang}, \binits{W.}},
\bauthor{\bsnm{Wang}, \binits{W.}},
\bauthor{\bsnm{Yang}, \binits{H.}},
\bauthor{\bsnm{Li}, \binits{P.}}:
\bctitle{Dstagnn: Dynamic spatial-temporal aware graph neural network for traffic flow forecasting}.
In: \bbtitle{International Conference on Machine Learning},
pp. \bfpage{11906}--\blpage{11917}
(\byear{2022}).
\bcomment{PMLR}
\end{bchapter}
\endbibitem

\bibitem[\protect\citeauthoryear{Cirstea et~al.}{2022}]{STWA}
\begin{botherref}
\oauthor{\bsnm{Cirstea}, \binits{R.-G.}},
\oauthor{\bsnm{Yang}, \binits{B.}},
\oauthor{\bsnm{Guo}, \binits{C.}},
\oauthor{\bsnm{Kieu}, \binits{T.}},
\oauthor{\bsnm{Pan}, \binits{S.}}:
Towards Spatio-Temporal Aware Traffic Time Series Forecasting--Full Version.
arXiv
(2022).
\doiurl{10.48550/ARXIV.2203.15737} .
\url{https://arxiv.org/abs/2203.15737}
\end{botherref}
\endbibitem

\bibitem[\protect\citeauthoryear{Zivot and Wang}{2006}]{zivot2006vector}
\begin{botherref}
\oauthor{\bsnm{Zivot}, \binits{E.}},
\oauthor{\bsnm{Wang}, \binits{J.}}:
Vector autoregressive models for multivariate time series.
Modeling financial time series with S-PLUS{\textregistered},
385--429
(2006)
\end{botherref}
\endbibitem

\bibitem[\protect\citeauthoryear{Vanajakshi and Rilett}{2004}]{10}
\begin{bchapter}
\bauthor{\bsnm{Vanajakshi}, \binits{L.}},
\bauthor{\bsnm{Rilett}, \binits{L.R.}}:
\bctitle{A comparison of the performance of artificial neural networks and support vector machines for the prediction of traffic speed}.
In: \bbtitle{Intelligent Vehicles Symposium}
(\byear{2004})
\end{bchapter}
\endbibitem

\bibitem[\protect\citeauthoryear{Dong et~al.}{2018}]{dong2018short}
\begin{bchapter}
\bauthor{\bsnm{Dong}, \binits{X.}},
\bauthor{\bsnm{Lei}, \binits{T.}},
\bauthor{\bsnm{Jin}, \binits{S.}},
\bauthor{\bsnm{Hou}, \binits{Z.}}:
\bctitle{Short-term traffic flow prediction based on xgboost}.
In: \bbtitle{2018 IEEE 7th Data Driven Control and Learning Systems Conference (DDCLS)},
pp. \bfpage{854}--\blpage{859}
(\byear{2018}).
\bcomment{IEEE}
\end{bchapter}
\endbibitem

\bibitem[\protect\citeauthoryear{Hochreiter and Schmidhuber}{1997}]{hochreiter1997long}
\begin{barticle}
\bauthor{\bsnm{Hochreiter}, \binits{S.}},
\bauthor{\bsnm{Schmidhuber}, \binits{J.}}:
\batitle{Long short-term memory}.
\bjtitle{Neural computation}
\bvolume{9}(\bissue{8}),
\bfpage{1735}--\blpage{1780}
(\byear{1997})
\end{barticle}
\endbibitem

\bibitem[\protect\citeauthoryear{Chung et~al.}{2014}]{chung2014empirical}
\begin{botherref}
\oauthor{\bsnm{Chung}, \binits{J.}},
\oauthor{\bsnm{Gulcehre}, \binits{C.}},
\oauthor{\bsnm{Cho}, \binits{K.}},
\oauthor{\bsnm{Bengio}, \binits{Y.}}:
Empirical evaluation of gated recurrent neural networks on sequence modeling.
arXiv preprint arXiv:1412.3555
(2014)
\end{botherref}
\endbibitem

\bibitem[\protect\citeauthoryear{Sutskever et~al.}{2014}]{sutskever2014sequence}
\begin{botherref}
\oauthor{\bsnm{Sutskever}, \binits{I.}},
\oauthor{\bsnm{Vinyals}, \binits{O.}},
\oauthor{\bsnm{Le}, \binits{Q.V.}}:
Sequence to sequence learning with neural networks.
Advances in neural information processing systems
\textbf{27}
(2014)
\end{botherref}
\endbibitem

\bibitem[\protect\citeauthoryear{Shi et~al.}{2015}]{shi2015convolutional}
\begin{botherref}
\oauthor{\bsnm{Shi}, \binits{X.}},
\oauthor{\bsnm{Chen}, \binits{Z.}},
\oauthor{\bsnm{Wang}, \binits{H.}},
\oauthor{\bsnm{Yeung}, \binits{D.-Y.}},
\oauthor{\bsnm{Wong}, \binits{W.-K.}},
\oauthor{\bsnm{Woo}, \binits{W.-c.}}:
Convolutional lstm network: A machine learning approach for precipitation nowcasting.
Advances in neural information processing systems
\textbf{28}
(2015)
\end{botherref}
\endbibitem

\bibitem[\protect\citeauthoryear{Zhang et~al.}{2016}]{zhang2016dnn}
\begin{bchapter}
\bauthor{\bsnm{Zhang}, \binits{J.}},
\bauthor{\bsnm{Zheng}, \binits{Y.}},
\bauthor{\bsnm{Qi}, \binits{D.}},
\bauthor{\bsnm{Li}, \binits{R.}},
\bauthor{\bsnm{Yi}, \binits{X.}}:
\bctitle{Dnn-based prediction model for spatio-temporal data}.
In: \bbtitle{Proceedings of the 24th ACM SIGSPATIAL International Conference on Advances in Geographic Information Systems},
pp. \bfpage{1}--\blpage{4}
(\byear{2016})
\end{bchapter}
\endbibitem

\bibitem[\protect\citeauthoryear{Zhang et~al.}{2017}]{zhang2017deep}
\begin{bchapter}
\bauthor{\bsnm{Zhang}, \binits{J.}},
\bauthor{\bsnm{Zheng}, \binits{Y.}},
\bauthor{\bsnm{Qi}, \binits{D.}}:
\bctitle{Deep spatio-temporal residual networks for citywide crowd flows prediction}.
In: \bbtitle{Thirty-first AAAI Conference on Artificial Intelligence}
(\byear{2017})
\end{bchapter}
\endbibitem

\bibitem[\protect\citeauthoryear{Yao et~al.}{2019}]{yao2019revisiting}
\begin{bchapter}
\bauthor{\bsnm{Yao}, \binits{H.}},
\bauthor{\bsnm{Tang}, \binits{X.}},
\bauthor{\bsnm{Wei}, \binits{H.}},
\bauthor{\bsnm{Zheng}, \binits{G.}},
\bauthor{\bsnm{Li}, \binits{Z.}}:
\bctitle{Revisiting spatial-temporal similarity: A deep learning framework for traffic prediction}.
In: \bbtitle{Proceedings of the AAAI Conference on Artificial Intelligence},
vol. \bseriesno{33},
pp. \bfpage{5668}--\blpage{5675}
(\byear{2019})
\end{bchapter}
\endbibitem

\bibitem[\protect\citeauthoryear{Fang et~al.}{2023}]{fang2023spatio}
\begin{bchapter}
\bauthor{\bsnm{Fang}, \binits{Y.}},
\bauthor{\bsnm{Qin}, \binits{Y.}},
\bauthor{\bsnm{Luo}, \binits{H.}},
\bauthor{\bsnm{Zhao}, \binits{F.}},
\bauthor{\bsnm{Xu}, \binits{B.}},
\bauthor{\bsnm{Zeng}, \binits{L.}},
\bauthor{\bsnm{Wang}, \binits{C.}}:
\bctitle{When spatio-temporal meet wavelets: Disentangled traffic forecasting via efficient spectral graph attention networks}.
In: \bbtitle{2023 IEEE 39th International Conference on Data Engineering (ICDE)},
pp. \bfpage{517}--\blpage{529}
(\byear{2023}).
\bcomment{IEEE}
\end{bchapter}
\endbibitem

\bibitem[\protect\citeauthoryear{Zhou et~al.}{2021}]{zhou2021informer}
\begin{bchapter}
\bauthor{\bsnm{Zhou}, \binits{H.}},
\bauthor{\bsnm{Zhang}, \binits{S.}},
\bauthor{\bsnm{Peng}, \binits{J.}},
\bauthor{\bsnm{Zhang}, \binits{S.}},
\bauthor{\bsnm{Li}, \binits{J.}},
\bauthor{\bsnm{Xiong}, \binits{H.}},
\bauthor{\bsnm{Zhang}, \binits{W.}}:
\bctitle{Informer: Beyond efficient transformer for long sequence time-series forecasting}.
In: \bbtitle{Proceedings of the AAAI Conference on Artificial Intelligence},
vol. \bseriesno{35},
pp. \bfpage{11106}--\blpage{11115}
(\byear{2021})
\end{bchapter}
\endbibitem

\bibitem[\protect\citeauthoryear{Goldberger et~al.}{2003}]{goldberger2003efficient}
\begin{bchapter}
\bauthor{\bsnm{Goldberger}, \binits{J.}},
\bauthor{\bsnm{Gordon}, \binits{S.}},
\bauthor{\bsnm{Greenspan}, \binits{H.}}, \betal:
\bctitle{An efficient image similarity measure based on approximations of kl-divergence between two gaussian mixtures.}
In: \bbtitle{ICCV},
vol. \bseriesno{3},
pp. \bfpage{487}--\blpage{493}
(\byear{2003})
\end{bchapter}
\endbibitem

\bibitem[\protect\citeauthoryear{Kitaev et~al.}{2020}]{kitaev2020reformer}
\begin{botherref}
\oauthor{\bsnm{Kitaev}, \binits{N.}},
\oauthor{\bsnm{Kaiser}, \binits{{\L}.}},
\oauthor{\bsnm{Levskaya}, \binits{A.}}:
Reformer: The efficient transformer.
arXiv preprint arXiv:2001.04451
(2020)
\end{botherref}
\endbibitem

\bibitem[\protect\citeauthoryear{Zhou et~al.}{2022}]{pmlr-v162-zhou22g}
\begin{bchapter}
\bauthor{\bsnm{Zhou}, \binits{T.}},
\bauthor{\bsnm{Ma}, \binits{Z.}},
\bauthor{\bsnm{Wen}, \binits{Q.}},
\bauthor{\bsnm{Wang}, \binits{X.}},
\bauthor{\bsnm{Sun}, \binits{L.}},
\bauthor{\bsnm{Jin}, \binits{R.}}:
\bctitle{{FED}former: Frequency enhanced decomposed transformer for long-term series forecasting}.
In: \bbtitle{Proceedings of the 39th International Conference on Machine Learning}.
\bsertitle{Proceedings of Machine Learning Research},
vol. \bseriesno{162},
pp. \bfpage{27268}--\blpage{27286}.
\bpublisher{PMLR}, \blocation{???}
(\byear{2022})
\end{bchapter}
\endbibitem

\bibitem[\protect\citeauthoryear{Jin et~al.}{2023}]{jin2023trafformer}
\begin{bchapter}
\bauthor{\bsnm{Jin}, \binits{D.}},
\bauthor{\bsnm{Shi}, \binits{J.}},
\bauthor{\bsnm{Wang}, \binits{R.}},
\bauthor{\bsnm{Li}, \binits{Y.}},
\bauthor{\bsnm{Huang}, \binits{Y.}},
\bauthor{\bsnm{Yang}, \binits{Y.-B.}}:
\bctitle{Trafformer: Unify time and space in traffic prediction}.
In: \bbtitle{Proceedings of the AAAI Conference on Artificial Intelligence},
vol. \bseriesno{37},
pp. \bfpage{8114}--\blpage{8122}
(\byear{2023})
\end{bchapter}
\endbibitem

\bibitem[\protect\citeauthoryear{Bruna et~al.}{2013}]{bruna2013spectral}
\begin{botherref}
\oauthor{\bsnm{Bruna}, \binits{J.}},
\oauthor{\bsnm{Zaremba}, \binits{W.}},
\oauthor{\bsnm{Szlam}, \binits{A.}},
\oauthor{\bsnm{LeCun}, \binits{Y.}}:
Spectral networks and locally connected networks on graphs.
arXiv preprint arXiv:1312.6203
(2013)
\end{botherref}
\endbibitem

\bibitem[\protect\citeauthoryear{Defferrard et~al.}{2016}]{defferrard2016convolutional}
\begin{botherref}
\oauthor{\bsnm{Defferrard}, \binits{M.}},
\oauthor{\bsnm{Bresson}, \binits{X.}},
\oauthor{\bsnm{Vandergheynst}, \binits{P.}}:
Convolutional neural networks on graphs with fast localized spectral filtering.
Advances in neural information processing systems
\textbf{29}
(2016)
\end{botherref}
\endbibitem

\bibitem[\protect\citeauthoryear{Micheli}{2009}]{micheli2009neural}
\begin{barticle}
\bauthor{\bsnm{Micheli}, \binits{A.}}:
\batitle{Neural network for graphs: A contextual constructive approach}.
\bjtitle{IEEE Transactions on Neural Networks}
\bvolume{20}(\bissue{3}),
\bfpage{498}--\blpage{511}
(\byear{2009})
\end{barticle}
\endbibitem

\bibitem[\protect\citeauthoryear{Song et~al.}{2020}]{song2020spatial}
\begin{bchapter}
\bauthor{\bsnm{Song}, \binits{C.}},
\bauthor{\bsnm{Lin}, \binits{Y.}},
\bauthor{\bsnm{Guo}, \binits{S.}},
\bauthor{\bsnm{Wan}, \binits{H.}}:
\bctitle{Spatial-temporal synchronous graph convolutional networks: A new framework for spatial-temporal network data forecasting}.
In: \bbtitle{Proceedings of the AAAI Conference on Artificial Intelligence},
vol. \bseriesno{34},
pp. \bfpage{914}--\blpage{921}
(\byear{2020})
\end{bchapter}
\endbibitem

\bibitem[\protect\citeauthoryear{Velickovic et~al.}{2017}]{velickovic2017graph}
\begin{barticle}
\bauthor{\bsnm{Velickovic}, \binits{P.}},
\bauthor{\bsnm{Cucurull}, \binits{G.}},
\bauthor{\bsnm{Casanova}, \binits{A.}},
\bauthor{\bsnm{Romero}, \binits{A.}},
\bauthor{\bsnm{Lio}, \binits{P.}},
\bauthor{\bsnm{Bengio}, \binits{Y.}}:
\batitle{Graph attention networks}.
\bjtitle{stat}
\bvolume{1050},
\bfpage{20}
(\byear{2017})
\end{barticle}
\endbibitem

\bibitem[\protect\citeauthoryear{Guo and Yuan}{2020}]{guo2020short}
\begin{barticle}
\bauthor{\bsnm{Guo}, \binits{G.}},
\bauthor{\bsnm{Yuan}, \binits{W.}}:
\batitle{Short-term traffic speed forecasting based on graph attention temporal convolutional networks}.
\bjtitle{Neurocomputing}
\bvolume{410},
\bfpage{387}--\blpage{393}
(\byear{2020})
\end{barticle}
\endbibitem

\bibitem[\protect\citeauthoryear{Zhang et~al.}{2019}]{zhang2019spatial}
\begin{barticle}
\bauthor{\bsnm{Zhang}, \binits{C.}},
\bauthor{\bsnm{James}, \binits{J.}},
\bauthor{\bsnm{Liu}, \binits{Y.}}:
\batitle{Spatial-temporal graph attention networks: A deep learning approach for traffic forecasting}.
\bjtitle{IEEE Access}
\bvolume{7},
\bfpage{166246}--\blpage{166256}
(\byear{2019})
\end{barticle}
\endbibitem

\bibitem[\protect\citeauthoryear{Huang et~al.}{2022}]{huang2022learning}
\begin{botherref}
\oauthor{\bsnm{Huang}, \binits{J.}},
\oauthor{\bsnm{Luo}, \binits{K.}},
\oauthor{\bsnm{Cao}, \binits{L.}},
\oauthor{\bsnm{Wen}, \binits{Y.}},
\oauthor{\bsnm{Zhong}, \binits{S.}}:
Learning multiaspect traffic couplings by multirelational graph attention networks for traffic prediction.
IEEE Transactions on Intelligent Transportation Systems
(2022)
\end{botherref}
\endbibitem

\bibitem[\protect\citeauthoryear{Li and Zhu}{2021}]{li2021spatial}
\begin{bchapter}
\bauthor{\bsnm{Li}, \binits{M.}},
\bauthor{\bsnm{Zhu}, \binits{Z.}}:
\bctitle{Spatial-temporal fusion graph neural networks for traffic flow forecasting}.
In: \bbtitle{Proceedings of the AAAI Conference on Artificial Intelligence},
vol. \bseriesno{35},
pp. \bfpage{4189}--\blpage{4196}
(\byear{2021})
\end{bchapter}
\endbibitem

\bibitem[\protect\citeauthoryear{Ye et~al.}{2022}]{ye2022learning}
\begin{bchapter}
\bauthor{\bsnm{Ye}, \binits{J.}},
\bauthor{\bsnm{Liu}, \binits{Z.}},
\bauthor{\bsnm{Du}, \binits{B.}},
\bauthor{\bsnm{Sun}, \binits{L.}},
\bauthor{\bsnm{Li}, \binits{W.}},
\bauthor{\bsnm{Fu}, \binits{Y.}},
\bauthor{\bsnm{Xiong}, \binits{H.}}:
\bctitle{Learning the evolutionary and multi-scale graph structure for multivariate time series forecasting}.
In: \bbtitle{Proceedings of the 28th ACM SIGKDD Conference on Knowledge Discovery and Data Mining},
pp. \bfpage{2296}--\blpage{2306}
(\byear{2022})
\end{bchapter}
\endbibitem

\bibitem[\protect\citeauthoryear{Elsayed et~al.}{2021}]{elsayed2021we}
\begin{botherref}
\oauthor{\bsnm{Elsayed}, \binits{S.}},
\oauthor{\bsnm{Thyssens}, \binits{D.}},
\oauthor{\bsnm{Rashed}, \binits{A.}},
\oauthor{\bsnm{Jomaa}, \binits{H.S.}},
\oauthor{\bsnm{Schmidt-Thieme}, \binits{L.}}:
Do we really need deep learning models for time series forecasting?
arXiv preprint arXiv:2101.02118
(2021)
\end{botherref}
\endbibitem

\bibitem[\protect\citeauthoryear{Wu et~al.}{2019}]{wu2019graph}
\begin{botherref}
\oauthor{\bsnm{Wu}, \binits{Z.}},
\oauthor{\bsnm{Pan}, \binits{S.}},
\oauthor{\bsnm{Long}, \binits{G.}},
\oauthor{\bsnm{Jiang}, \binits{J.}},
\oauthor{\bsnm{Zhang}, \binits{C.}}:
Graph wavenet for deep spatial-temporal graph modeling.
arXiv preprint arXiv:1906.00121
(2019)
\end{botherref}
\endbibitem

\bibitem[\protect\citeauthoryear{Bai et~al.}{2020}]{bai2020adaptive}
\begin{barticle}
\bauthor{\bsnm{Bai}, \binits{L.}},
\bauthor{\bsnm{Yao}, \binits{L.}},
\bauthor{\bsnm{Li}, \binits{C.}},
\bauthor{\bsnm{Wang}, \binits{X.}},
\bauthor{\bsnm{Wang}, \binits{C.}}:
\batitle{Adaptive graph convolutional recurrent network for traffic forecasting}.
\bjtitle{Advances in neural information processing systems}
\bvolume{33},
\bfpage{17804}--\blpage{17815}
(\byear{2020})
\end{barticle}
\endbibitem

\bibitem[\protect\citeauthoryear{Irie and Miyake}{1988}]{irie1988capabilities}
\begin{bchapter}
\bauthor{\bsnm{Irie}, \binits{B.}},
\bauthor{\bsnm{Miyake}, \binits{S.}}:
\bctitle{Capabilities of three-layered perceptrons.}
In: \bbtitle{ICNN},
pp. \bfpage{641}--\blpage{648}
(\byear{1988})
\end{bchapter}
\endbibitem

\bibitem[\protect\citeauthoryear{Srivastava et~al.}{2014}]{srivastava2014dropout}
\begin{barticle}
\bauthor{\bsnm{Srivastava}, \binits{N.}},
\bauthor{\bsnm{Hinton}, \binits{G.}},
\bauthor{\bsnm{Krizhevsky}, \binits{A.}},
\bauthor{\bsnm{Sutskever}, \binits{I.}},
\bauthor{\bsnm{Salakhutdinov}, \binits{R.}}:
\batitle{Dropout: a simple way to prevent neural networks from overfitting}.
\bjtitle{The journal of machine learning research}
\bvolume{15}(\bissue{1}),
\bfpage{1929}--\blpage{1958}
(\byear{2014})
\end{barticle}
\endbibitem

\bibitem[\protect\citeauthoryear{Zhao et~al.}{2023}]{zhao2023multi}
\begin{barticle}
\bauthor{\bsnm{Zhao}, \binits{W.}},
\bauthor{\bsnm{Zhang}, \binits{S.}},
\bauthor{\bsnm{Zhou}, \binits{B.}},
\bauthor{\bsnm{Wang}, \binits{B.}}:
\batitle{Multi-spatio-temporal fusion graph recurrent network for traffic forecasting}.
\bjtitle{Engineering Applications of Artificial Intelligence}
\bvolume{124},
\bfpage{106615}
(\byear{2023})
\end{barticle}
\endbibitem

\bibitem[\protect\citeauthoryear{Paszke et~al.}{2019}]{paszke2019pytorch}
\begin{botherref}
\oauthor{\bsnm{Paszke}, \binits{A.}},
\oauthor{\bsnm{Gross}, \binits{S.}},
\oauthor{\bsnm{Massa}, \binits{F.}},
\oauthor{\bsnm{Lerer}, \binits{A.}},
\oauthor{\bsnm{Bradbury}, \binits{J.}},
\oauthor{\bsnm{Chanan}, \binits{G.}},
\oauthor{\bsnm{Killeen}, \binits{T.}},
\oauthor{\bsnm{Lin}, \binits{Z.}},
\oauthor{\bsnm{Gimelshein}, \binits{N.}},
\oauthor{\bsnm{Antiga}, \binits{L.}}, et al.:
Pytorch: An imperative style, high-performance deep learning library.
Advances in neural information processing systems
\textbf{32}
(2019)
\end{botherref}
\endbibitem

\end{thebibliography}

\end{document}